%% file: paper.tex
\documentclass[letterpaper, 10 pt, journal, twoside]{ieeetran}

\IEEEoverridecommandlockouts                              % This command is only
\input{definitions_drs}

\input{definitions_custom}

\newcommand{\papertitle}[1]{
	An Efficient Locally Reactive Controller for Safe Navigation in Visual 
	Teach and Repeat Missions
}
\newcommand{\shorttitle}[1]{
	An Efficient Locally Reactive Controller for Safe Visual 
	Teach and Repeat
}

\newcommand{\surname}[1]{Mattamala}

\newcommand{\videolink}{https://youtu.be/G_AwNec5AwU}

\title {An Efficient Locally Reactive Controller for Safe Navigation in Visual 
Teach and Repeat Missions}

\author{
	Matias Mattamala$^{1}$,
	Nived Chebrolu$^{1}$,
	and Maurice Fallon$^{1}$
	\thanks{Manuscript received: September, 9, 2021; Revised November, 25, 
	2021; Accepted December, 29, 2021.}%Use only for final RAL version
	\thanks{This paper was recommended for publication by Editor Eric Marchand 
	upon evaluation of the Associate Editor and Reviewers' comments.
	This work is supported by the ESPRC/UKRI ORCA Robotics Hub
	(EP/R026173/1), a Royal Society University Research Fellowship (Fallon),
	and the ANID~/~Scholarship Program~/~DOCTORADO BECAS 
	CHILE~/~2019-72200291 (Mattamala).
	This research has been conducted as part of the ANYmal research 
	community.} %Use only for final RAL version

	\thanks{$^{1}$The authors are with the Oxford Robotics Institute at the 
	University of Oxford, UK. 
	{\tt\footnotesize \{matias, nived, mfallon\}@robots.ox.ac.uk}. 
	}
	\thanks{Digital Object Identifier (DOI): see top of this page.}
}

\begin{document}
\bstctlcite{library:BSTcontrol}

%%%%%%%%%%%%%%%%%%%%%%%%%%%%%%%%%%%%%%%%%%%%%%%%%%%%%%%%%%%%%%%%%%%%%%%%%%%%%%%%
% Title
\maketitle

%%%%%%%%%%%%%%%%%%%%%%%%%%%%%%%%%%%%%%%%%%%%%%%%%%%%%%%%%%%%%%%%%%%%%%%%%%%%%%%%
% ABSTRACT

\begin{abstract}
To achieve successful field autonomy, mobile robots need to 
freely adapt to changes in their environment. Visual navigation systems such as 
Visual Teach and Repeat (\vtr{}) often assume the space around the reference 
trajectory is free, but if the environment is obstructed 
path tracking can fail or the robot could collide with a previously unseen obstacle. In this work, we present a 
locally reactive controller for a \vtr{} system that allows a robot to navigate 
safely despite physical changes to the environment. Our controller uses a local elevation map to compute vector representations 
and outputs twist commands for navigation at \SI{10}{\hertz}. They are combined in a Riemannian Motion 
Policies (RMP) controller that requires $<$~\SI{2}{\milli\second} to run on a 
CPU. We integrated our controller with a \vtr{} system onboard an ANYmal C 
robot and tested it in indoor cluttered spaces and a large-scale 
underground mine. We demonstrate that our locally reactive controller keeps the 
robot safe when physical occlusions or loss of visual tracking occur such as 
when walking close to walls, crossing doorways, or traversing narrow corridors.
\end{abstract}

%%%%%%%%%%%%%%%%%%%%%%%%%%%%%%%%%%%%%%%%%%%%%%%%%%%%%%%%%%%%%%%%%%%%%%%%%%%%%%%%
% Keywords

%%%%%%%%%%%%%%%%%%%%%%%%%%%%%%%%%%%%%%%%%%%%%%%%%%%%%%%%%%%%%%%%%%%%%%%%%%%%%%%%
%\section*{Supplementary Material}
%\todo[inline, color=yellow]{Add links to videos, code, etc}

%%%%%%%%%%%%%%%%%%%%%%%%%%%%%%%%%%%%%%%%%%%%%%%%%%%%%%%%%%%%%%%%%%%%%%%%%%%%%%%%
\section{Introduction}
\label{sec:introduction}

\IEEEPARstart{S}{afe} navigation is a fundamental capability required to 
successfully deploy 
robots in natural and human-made environments. Forests, mines, and industrial 
facilities are challenging places due to clutter, narrow passages, or moving 
objects that effect the configuration space in which a robot operates. Dealing 
with such difficulties requires a sense of local awareness, in order to 
effectively adapt the robot's behavior to the task and state of 
the environment.

Visual Teach and Repeat (\vtr{}) systems are a practical approach to enable
inspections or patrols of known places, or to navigate from 
point-to-point~\cite{Furgale2010, Fehr2018}. 
Instead of building a precise metric map, \vtr{} relies on a \emph{topo-metric} 
representation built during a \emph{teach} phase, which is then used during the 
\emph{repeat} phase to execute point-to-point missions using only camera input 
to track the reference path. However, in practice, the visual appearance of real environments can change (due to the lighting conditions or because of onboard illumination~\cite{Churchill2012}) as can the physical layout (such as when a pathway is 
blocked), risking the success of the mission.

In this paper, we present a locally reactive controller that works as a safety 
layer to help a \vtr{} system achieve robust operation after significant changes to the 
environment. It allows a robot to be aware of its 
surroundings using a local map built online. We argue that for \vtr{} missions, 
we do not need to rely on a global and a local planner to navigate these 
environments. Instead, we exploit the fact that the taught path effectively 
represents a \emph{global plan in free space}, only requiring a fast, reactive 
controller to track the path safely, which is the main focus of this work.

\begin{figure}[t]
	\centering
	\includegraphics[width=\columnwidth]{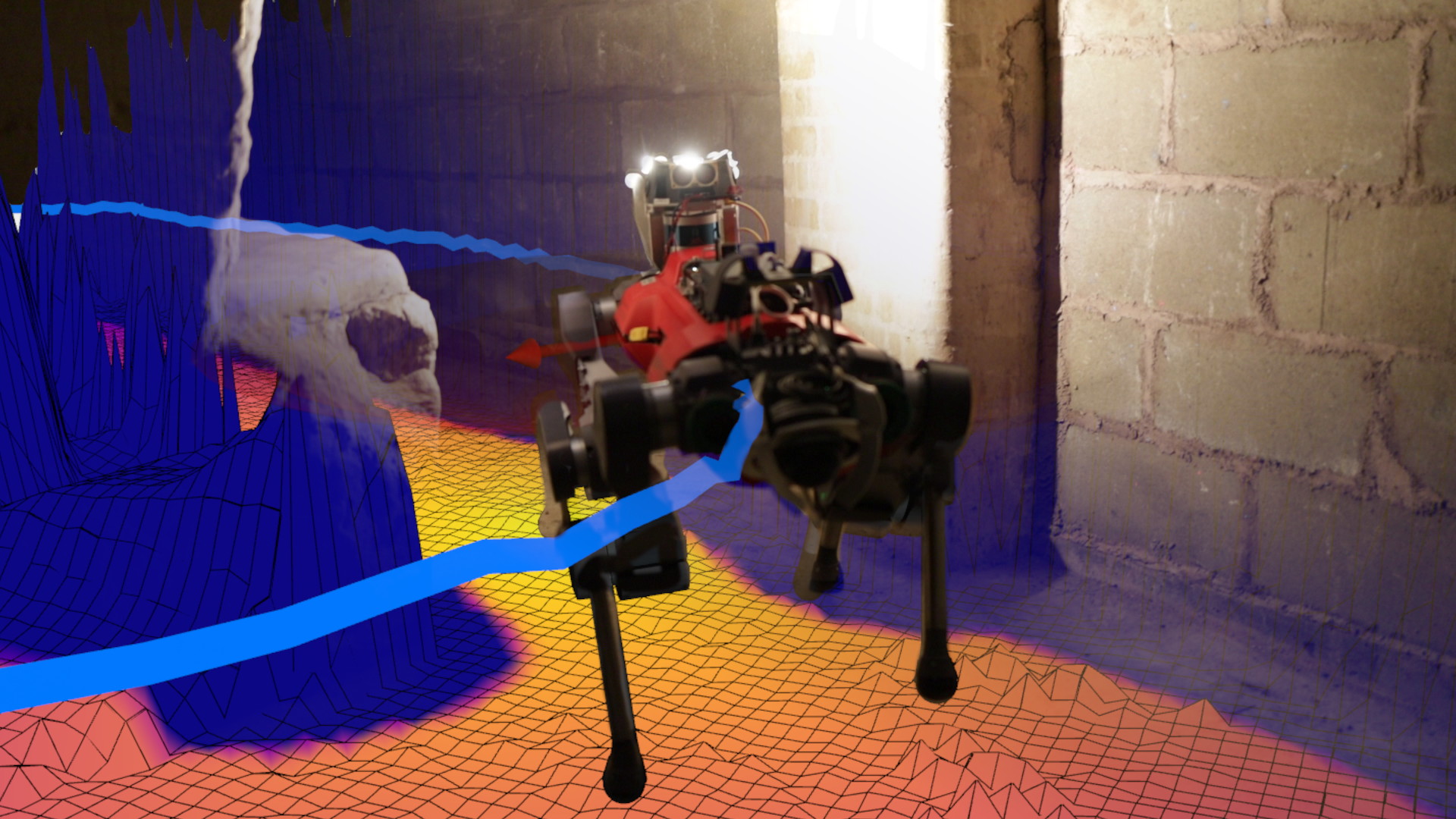}\vspace{1mm}
	\caption{\small{Our locally reactive controller allows an ANYmal 
			robot to safely execute Visual Teach and Repeat navigation 
			through narrow corridors in a decommissioned underground mine.
			We use the local map built on-the-fly to compute 
			efficient vector field representations, such as the \emph{geodesic 
			distance field} pictured, and combine them in a Riemannian 
			Motion Policies controller that runs at \SI{10}{\hertz}.\\
			Video: \texttt{\url{\videolink}}
	}}
	\label{fig:cover}
	\vspace{-2mm}
\end{figure}

Our controller uses the local map to generate efficient vector fields that 
provide the cues necessary for navigation, such as avoiding obstacles or 
reaching (local) goals around corners. The fields are combined in a principled 
manner using a Riemannian Motion Policy (RMP) formulation~\cite{Ratliff2018}, 
which allows us to dynamically enable specific behaviors depending on the 
robot's state. We demonstrate its closed-loop integration with a \vtr{} system, 
showing safe and fast navigation in cluttered indoor spaces and challenging 
large scale environments.

The main contributions of our work are the following:
\begin{itemize}
	\item A reactive RMP-based twist controller that uses a 
	local elevation map built online as the main perceptual input.
	\item Online computation of vector fields representations computed from the 
	local map at \SI{10}{\hertz} on a CPU.
	\item Closed-loop integration with our \vtr{} system~\cite{Mattamala2021} 
	on the ANYmal platform.
	\item Real-world validation using the ANYmal C quadruped in 
	realistic environments including scenarios with substantial change to the 
	scene and narrow pathways.
	\item First demonstration of a RMP reactive controller with a 
	full-sized dynamic robot in large-scale environments.
\end{itemize}

%%%%%%%%%%%%%%%%%%%%%%%%%%%%%%%%%%%%%%%%%%%%%%%%%%%%%%%%%%%%%%%%%%%%%%%%%%%%%%%%
\section{Related Work}

\subsection{Collision-free Navigation}
The navigation problem aims to determine an action (or set of 
actions) that allows a robot to reach a specific goal without colliding with obstacles. 
This is typically achieved through \emph{reactive controllers} or \emph{local 
planners}~\cite{Kappler2018}. Reactive controllers leverage the 
information available at a specific time and determine a single action to be 
directly executed by the robot. In contrast, local planners typically generate 
a collision-free trajectory by considering a short look-ahead 
and solving an optimization problem over that region.

For legged robots, determining the optimal action or trajectory can involve the 
computation of footsteps, which is known as \emph{whole-body control/planning}. 
Buchanan \etal~\cite{Buchanan2021} combine the whole-body planning problem with 
obstacle information in the form of a 3D signed distance field that is used for 
collision checking. This allows the robot to navigate through narrow spaces 
while planning individual footsteps through traversable regions.  
Similarly, Gaertner \etal~\cite{Gaertner2021} demonstrate collision-free 
locomotion by means of a Model Predictive Controller (MPC). 
These approaches couple the collision-free navigation task with footstep 
computation, thereby making them harder to generalize to different platforms.

In this work, we use a twist interface to control the legged 
robot's body velocity. It has been successfully demonstrated on several 
platforms such as ANYmal C~\cite{Hoeller2021}, Mini 
Cheetah~\cite{Kim2020}, and Vision60~\cite{Vasilopoulos2020}. Our controller is 
based on the Riemannian Motion Policies (RMP) 
framework~\cite{Ratliff2018}, a recent approach to robot motion generation. It 
allows us to generate reactive behaviors by combining different \emph{tasks} 
--such as obstacle avoidance or heading corrections-- depending on the robot's 
state. By formulating the problem as reactive twist-based control, we can 
easily make this local safety layer robot and gait independent.

\subsection{Representations for Local Navigation}
Classical navigation approaches typically consider only the geometric information of the 
environment, which can be either represented by occupancy 
maps~\cite{Bista2021}, 
voxels~\cite{Bradley2015}, point clouds~\cite{Krusi2017}, 
meshes~\cite{Brandao2020} or height 
maps~\cite{Fankhauser2016GridMapLibrary}.
Typically, a traversibility map is computed from these representations giving a 
measure of how safe a region is for robot navigation.

Semantics can often help to expand the notion of traversable space. Bradley 
\etal~\cite{Bradley2015} learn four types of terrain used for navigation in a forest environment with 
the LS3 quadruped. Nardi and Stachniss~\cite{Nardi2019} as well as Wellhausen 
\etal~\cite{Wellhausen2019} learn terrain 
properties in a self-supervised fashion which are then used as costs for a path 
planning algorithm that enables them to navigate safely. 

Representations such as Signed Distance Fields 
(SDFs)~\cite{Oleynikova2016} are particularly suitable for navigation, since 
they encode the distance to the obstacle surfaces in the environment. The SDFs 
are also differentiable and lend themselves to computation of a vector field, 
which can be seamlessly integrated in optimization-based planners or 
vector-based controllers. 
Another useful representation are the Geodesic Distance Fields (GDF) which 
capture the shortest path to goal from different positions in the 
map~\cite{Mainprice2020}.
These fields can be computed efficiently using the Fast Marching 
Method~\cite{Sethian1996} or the Heat Method~\cite{Crane2017} and 
have been used in the past for navigation~\cite{Valero-Gomez2013, Putz2021}.
In this work, we also take advantage of these representations and their vector 
fields to generate specific behaviors. However, we only compute them in the 
local map, which suffices for reactive control, efficiently 
exploiting the 2.5D geometry for online operation.

\subsection{Safe Navigation for \vtr{}}
In the context of \vtr{} systems, it is usually assumed that the path is 
collision-free and only a few works have addressed safe navigation explicitly. 
Ostafew \etal~\cite{Ostafew2016} formulate the collision-free navigation 
problem as a 
non-linear MPC controller with constraint satisfaction. The constraints are 
given by potential obstacles or localization limits, i.e, forcing the robot to 
stay  close to the reference path to avoid localization failures.
Berczi and Barfoot~\cite{Berczi2017} learn models of traversability over
multiple runs, which are used to detect unexpected obstacles when compared 
to the reference experience. 

Recently, Bista \etal~\cite{Bista2021} presented a 
topological navigation system for indoor environments which heuristically 
combined the output 
of a visual servoing system with the solution of 
a local plan computed in a 2D grid with A$^{\star}$. While we share some 
general ideas, the fundamental differences are that our \vtr{} system 
explicitly computes a 6 DOF pose --because legged robots effectively move in 3D 
space--, while it also presents a principled way to combine different desired 
behaviors within the same control framework.

\begin{figure}[t]
	\centering
	\includegraphics[width=\columnwidth]{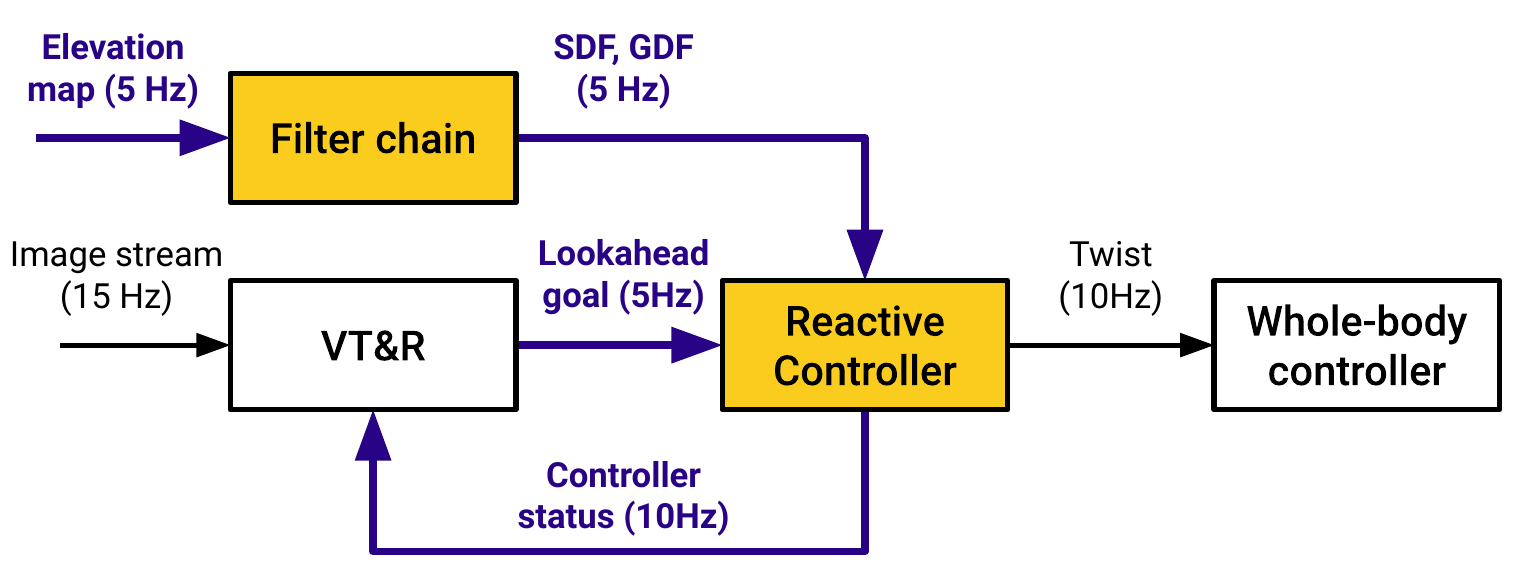}
	\caption{\small{Main pipeline of our safe \vtr{} system integrated with
			a quadruped robot. The \emph{filter chain} module computes vector 
			field representations from the elevation map 
			(\secref{sec:vector-representations}). They are used by the 
			\emph{reactive controller} module 
			(\secref{sec:locally-reactive-controller}) to 
			generate the command sent to the whole-body controller.}}
	\label{fig:pipeline}
\end{figure}

%%%%%%%%%%%%%%%%%%%%%%%%%%%%%%%%%%%%%%%%%%%%%%%%%%%%%%%%%%%%%%%%%%%%%%%%%%%%%%%%
\section{Method}
Our locally reactive controller and its integration with the \vtr{} system is 
summarized in \figref{fig:pipeline}. Our approach is a safety mechanism 
operating on the robot's local elevation map based on Riemannian Motion 
Policies (\secref{sec:preliminaries-rmp}). First, the \emph{Filter chain} 
module (\secref{sec:vector-representations}) determines the traversability of 
the local map and computes efficient vector field representations, such as SDFs 
and GDFs. These fields are fed as input to the \emph{Reactive Controller} 
module, which implements the RMP controller 
(\secref{sec:locally-reactive-controller}). Finally, we describe the 
integration of the reactive controller with our \vtr{} 
system in \secref{sec:vtr-integration}.

\subsection{Preliminaries: Riemannian Motion Policies}
\label{sec:preliminaries-rmp}

We base our controller on the Riemannian Motion Policies 
framework~\cite{Ratliff2018, Cheng2021}, which we 
briefly introduce. We consider 
a set of different desired behaviors or \emph{tasks} specified by tuples 
$\{\mathbf{f},\, \mathbf{M}\}$. Here $\mathbf{f} = \mathbf{f}(\mathbf{x}, 
\mathbf{\dot{x}})$ is an acceleration field or \emph{motion policy} that 
describes the behavior, namely goal reaching, obstacle avoidance, or damping. 
$\mathbf{M} = \mathbf{M}(\mathbf{x}, \mathbf{\dot{x}})$, is a matrix known as 
\emph{Riemannian metric}, which smoothly varies with the robot's state 
$\mathbf{x}$ and determines the importance of the field $\mathbf{f}$. 
The tuple is hence known as a \emph{Riemannian Motion Policy} (RMP).

This framework allows us to combine RMPs $\{\mathbf{f},\, 
\mathbf{M}\}$ from different sources to determine the optimal 
control $\mathbf{\ddot{x}}$ by minimizing:
\begin{equation}
\label{eq:rmp-formulation}
\argmin_{\mathbf{\ddot{x}}}{\displaystyle 
	\sum_{\{\mathbf{f},\, \mathbf{M} \}} 
	\lVert 
	\mathbf{f} - 
	\mathbf{\ddot{x}} 
	\rVert^{2}_{\mathbf{M}}}.
\end{equation}
Since the optimization is a linear least-squares problem, it can be solved in 
closed form, and the desired acceleration is the weighted average of the RMPs. 
The optimal $\mathbf{\ddot{x}}$ is forward integrated to obtain the twist 
command.

While the framework is general and some of its theoretical advantages have been 
shown~\cite{Cheng2021}, designing the different RMPs depends on the robot's 
sensors and the environment. In this work we are interested in exploiting 2.5D 
elevation maps commonly used by legged and other ground robots to represent 
the local environment, and in generating efficient vector field representations 
from them to design our RMPs.

\subsection{Efficient Vector Field Representations}
\label{sec:vector-representations}
\begin{figure*}[t]
	\vspace{1mm}
	\centering
	\includegraphics[trim={3cm 0cm 11cm 0cm},clip, 
	width=0.62\columnwidth]{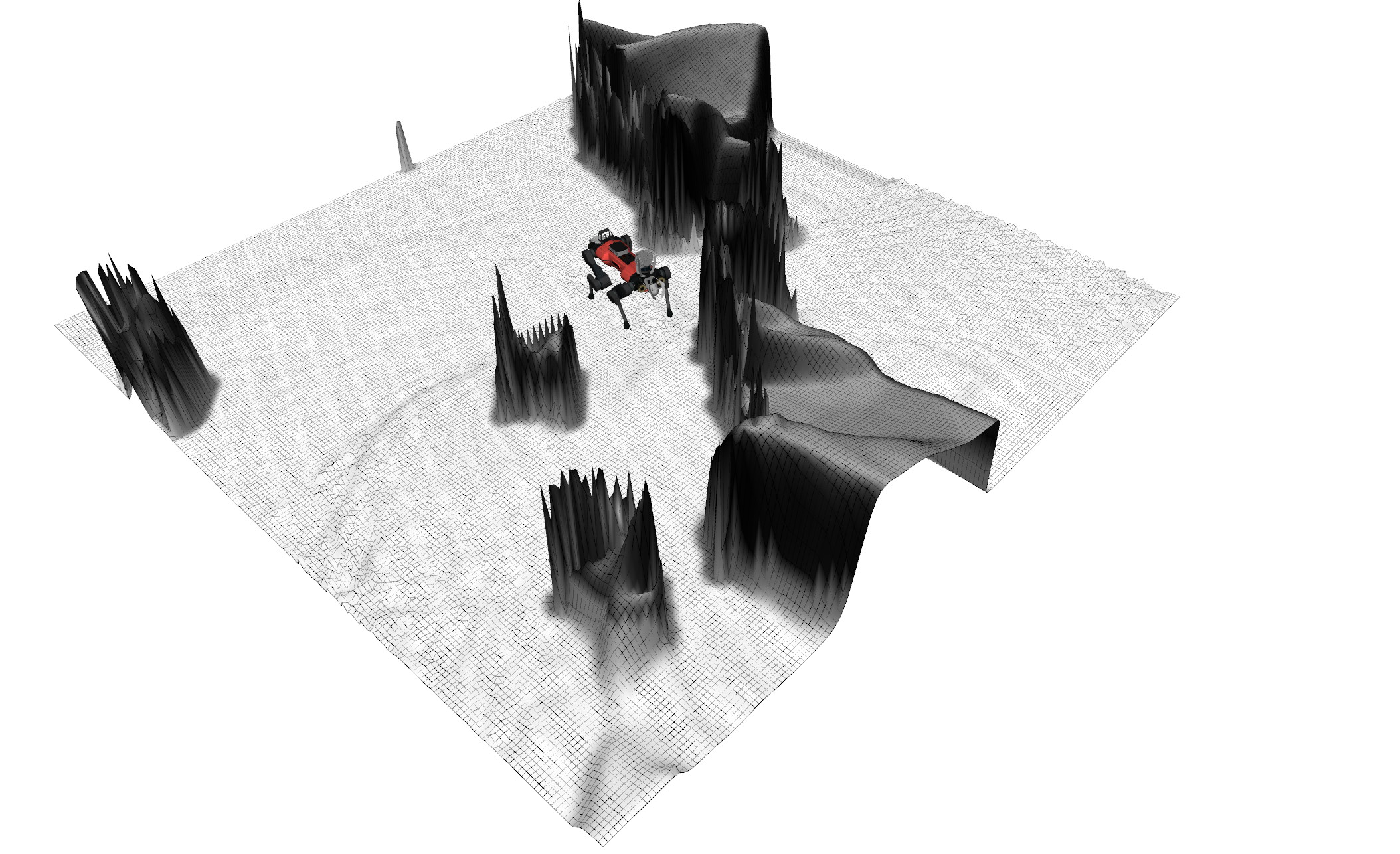}
	\includegraphics[trim={3cm 0cm 11cm 0cm},clip, 
	width=0.62\columnwidth]{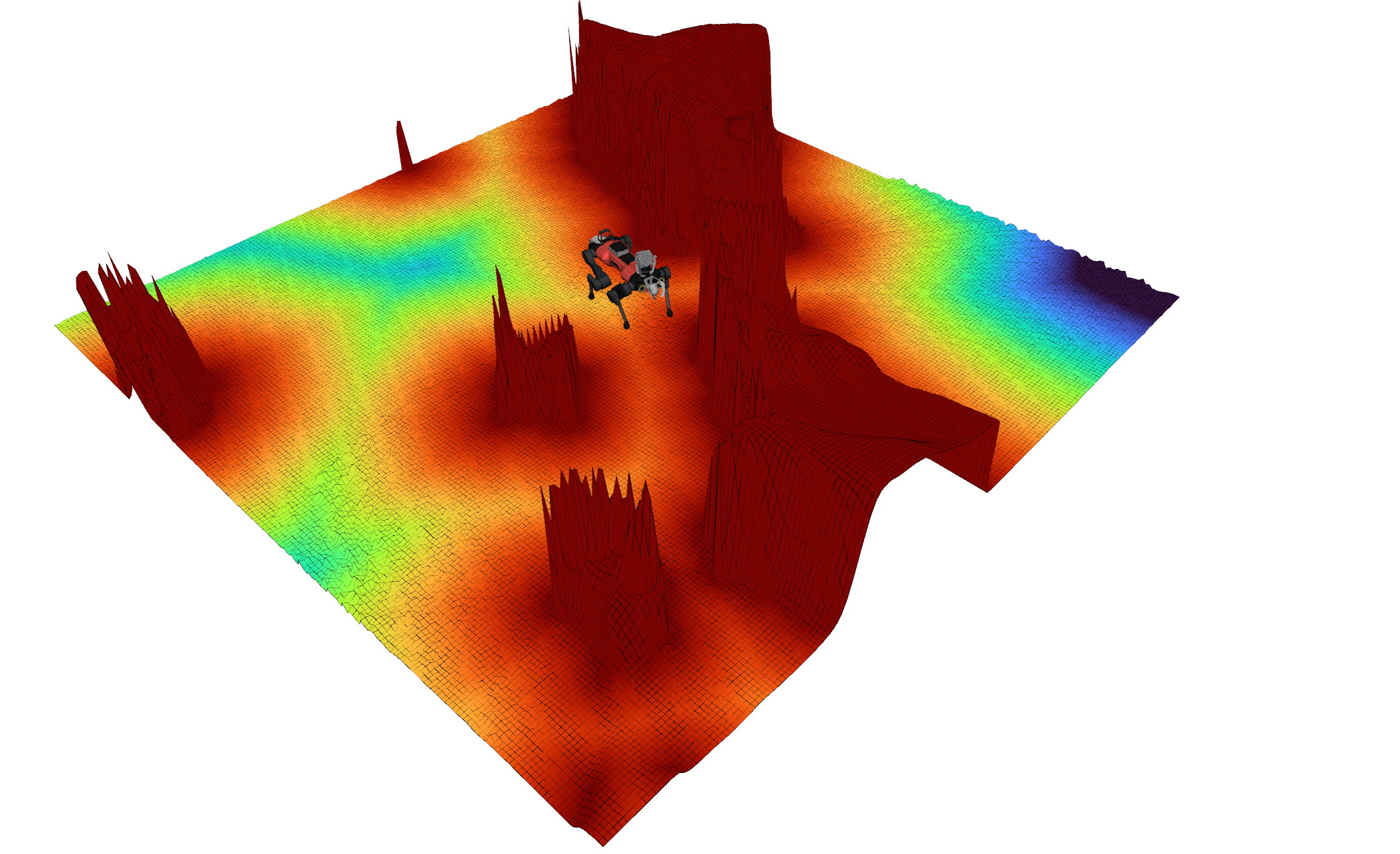}
	\includegraphics[trim={3cm 0cm 11cm 0cm},clip, 
	width=0.62\columnwidth]{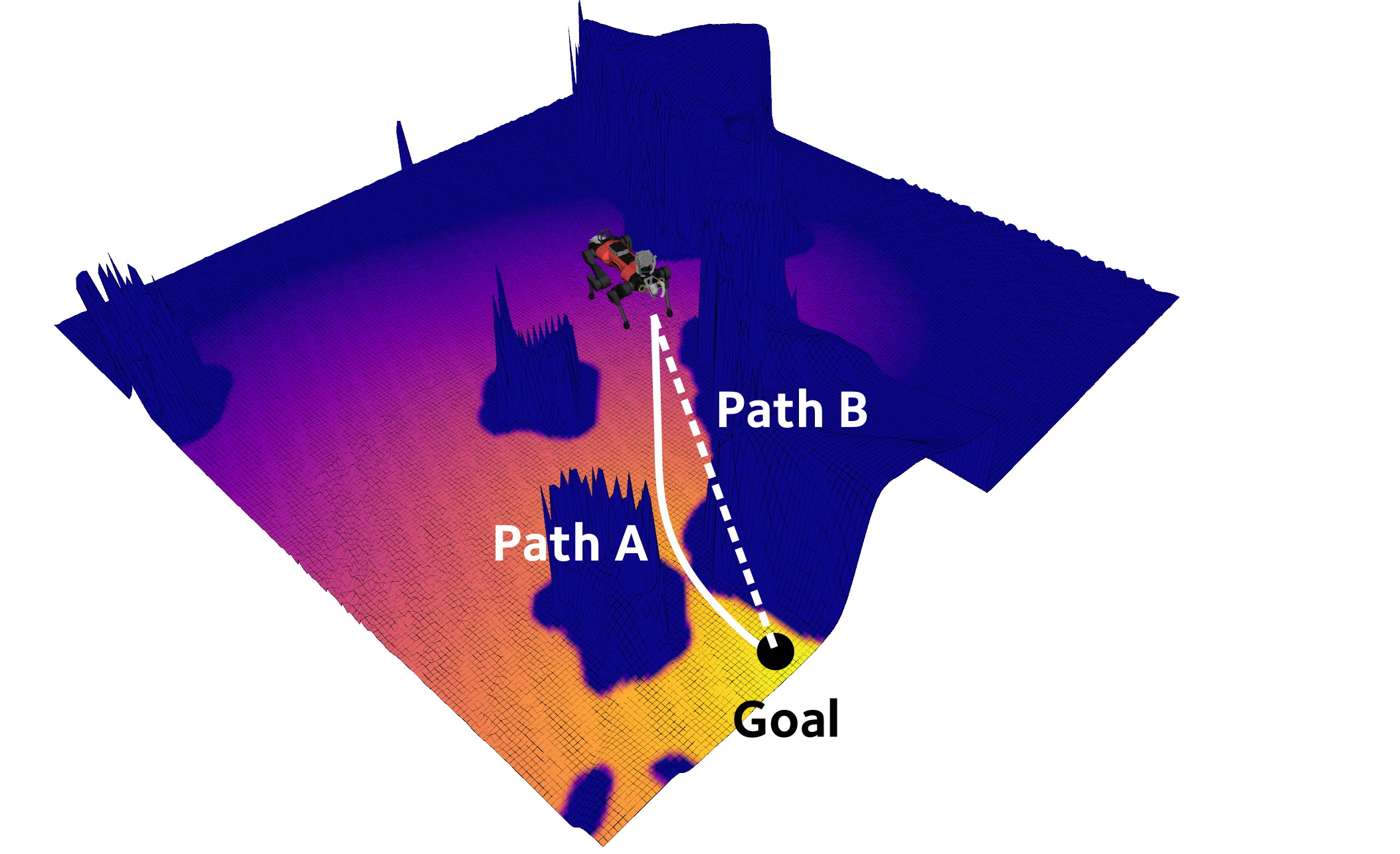}
	\caption{\small{Representations computed from the elevation map, as 
	described 
	in \secref{sec:vector-representations}.
	\textbf{Left:} Traversability layer computed from geometric analysis; light gray 
	means traversable, whereas black is not traversable.
	\textbf{Center:} Signed Distance Field 
	(SDF); color indicates the distance to obstacles from red (closer) to blue 
	(further). \textbf{Right:} Geodesic Distance Field (GDF); the goal is shown 
	in yellow, and further areas in dark blue. Using the GDF, path A has a 
	smaller cost compared to B, which goes over the obstacles.
	}}
	\label{fig:vector-representations}
	%\vspace{-8mm}
\end{figure*}

Elevation maps are a popular representation of the local environment
as they (1) preserve 2.5D information about the terrain, (2) can be 
easily computed from depth cameras or LiDARs via 
ray-casting~\cite{Fankhauser2018}, (3) they are now usually available in legged 
robots since they are used for locomotion tasks, and (4) they can be 
manipulated as images, making them suitable to 
efficient computer vision operations.

Thus, we assume that a local elevation map reconstructing a limited 
space around the robot is available. We base our implementation on the universal grid map 
library~\cite{Fankhauser2016GridMapLibrary}, which represents the local 
elevation map in a multi-layered fixed-sized grid. We use a \SI{8}{\meter} 
$\times$ \SI{8}{\meter} map with a resolution of \SI{4}{\centi\meter}, giving
a grid of $200\times200$ cells. To obtain the vector field representations, we 
follow the steps explained here:

\subsubsection{Filling unknown areas}
We first run an in-painting filter on the elevation map. 
Image-based in-painting methods fill the missing data using nearby pixels. We 
use the inpainting approach~\cite{Telea2004} implemented in OpenCV with a radius of 
\SI{10}{\centi\meter}. Please refer to \secref{sec:computation-time} for 
a further discussion.

\subsubsection{Traversability computation}
We compute a continuous measure for terrain traversability using the slope and roughness of the local map~\cite{Wermelinger2016}.
We additionally take into account the height of terrain with respect to the body pose. This means that if the robot is inclined, e.g. on a hill, the height is measured normal to the surface and
the sloped surface is treated as traversable. An example of the continuous 
traversability map is shown in 
\figref{fig:vector-representations} left. The continuous value is thresholded 
to produce a binary classification $t\in[0,1]$, where $1$ means fully 
traversable and $0$ is not traversable.

\subsubsection{Signed Distance Field (SDF)}
The SDF layer (\figref{fig:vector-representations} center) is computed using 
the 2D Euclidean distance transform~\cite{Felzenszwalb2012}. Given the binarized traversability layer, we 
apply the distance transform to obtain the field $f_{\text{SDF}}$. 
To compute the normalized gradient $\nabla f_{\text{SDF}}$, we use the Sobel 
edge-detector, which is further smoothed using image filters (Gaussian, 
median, etc). By relying on 2D image operations, we are able to compute the SDF 
online at \SI{10}{\hertz}, which is crucial to allow the controller to be 
responsive to new obstacles.

\subsubsection{Geodesic Distance Field (GDF)}
Geodesic distance fields are potential functions with an unique 
global minimum.
This is used to reach a goal from any point in a local map.
Similar to the SDF, we use the Fast Marching Method (FMM)~\cite{Sethian1996} on the binarized 
traversability layer to build the GDF layer. 
The FMM solves the Eikonal equation~\cite{Peyre2010} from a seed point, from 
which a wavefront is propagated to fill the entire space. In our 
implementation, the seed is given by the goal position in the local map or, if 
the goal falls outside the map, the closest point in the map is chosen. The 
resulting field is shown in \figref{fig:vector-representations} right. After 
$f_{\text{GDF}}$ is computed, we obtain normalized gradients 
using the Sobel operator, which define the \emph{geodesic flow} $\nabla 
f_{\text{GDF}}$ to reach the goal.

\subsection{Locally Reactive Controller}
\label{sec:locally-reactive-controller}
The reactive controller module in \figref{fig:pipeline} implements the 
RMP controller presented in \eqref{eq:rmp-formulation}, using the vector fields 
obtained from the elevation map to create goal reaching and obstacle avoidance 
RMPs. \tabref{tab:motion-policies} summarizes all the RMPs considered along the
expressions for the fields and metrics. However, 
there are aspects to consider for their implementation and integration 
with the navigation pipeline:

\subsubsection{State representation and reference frames} 
\begin{figure}[h]
	\centering
	\includegraphics[width=0.95\columnwidth]{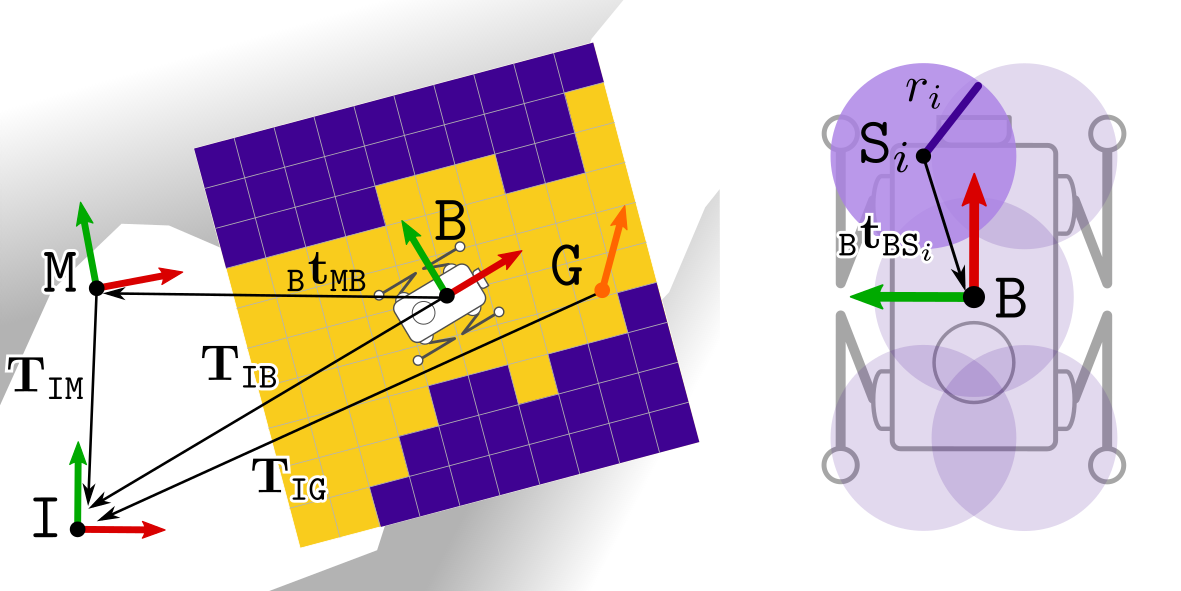}
	\caption{\small{Reference frames used in this work. \textbf{Left:} 
			The pose of the robot's 
			body $\notationFrame{B}$ in the inertial frame $\notationFrame{I}$ 
			is given by 
			$\notationMatrix{T}{\notationFrame{IB}}\in \SEtwo$. The local map, 
			expressed in frame $\notationFrame{B}$, represents a local 
			representation of the environment. \textbf{Right:} The collision 
			spheres $\mathcal{S}$ define the frame $\notationFrame{S}_i$ and 
			are used to represent the robot's volume, their positions are 
			expressed in the body frame by the vector 
			$\notationVectorFrame{t}{B}{\notationFrame{BS}_{i}}$.}}
	\label{fig:reference-frames}
\end{figure}

Our formulation considers the frame conventions illustrated in 
\figref{fig:reference-frames}. The state $\notatVector{x}$ of the robot is 
given by the pose of the robot's body 
$\notationMatrixFrame{T}{}{\notationFrame{IB}}\in\SEtwo$ expressed in the 
fixed, inertial frame $\notationFrame{I}$. This pose is given by our \vtr{} system or by 
other autonomy system. The robot's velocity 
$\notatVector{\dot{x}}$ and acceleration $\notatVector{\ddot{x}}$ are also 
expressed in the body frame $\notationFrame{B}$ and will be denoted onward by 
$\notationVectorFrame{v}{B}{} = (\notationScalarFrame{v}{B}{x}, 
\notationScalarFrame{v}{B}{y}, \notationScalarFrame{v}{B}{\theta}) \in 
\Rn{3}$ and $\notationVectorFrame{a}{B}{} = (\notationScalarFrame{a}{B}{x}, 
\notationScalarFrame{a}{B}{y}, \notationScalarFrame{a}{B}{\theta}) \in \Rn{3}$.

The goal pose is expressed in the inertial frame $\notationMatrix{T}{\notationFrame{IG}}\in\SEtwo$.  
The local map is computed in the map frame $\notationFrame{M}$ which has a translation offset of  $\notationVectorFrame{t}{B}{\notationFrame{MB}} \in 
\Rn{2}$ to coincide with the center of the robot body.

Note that the reference frame used by the motion policies are described in body frame~$\notationFrame{B}$,
whereas $\nabla f_{\text{SDF}}$ and $\nabla f_{\text{GDF}}$ are defined in the local map frame~$\notationFrame{M}$.
These gradients need to be properly transformed to $\notationFrame{B}$ before 
being integrated in the optimization problem in \eqref{eq:rmp-formulation}.
Please refer to the Appendix for technical details.

\subsubsection{Collision spheres}
To represent the robot's volume, we follow the same approach as previous works 
by using a set of collision spheres $\mathcal{S}$~\cite{Gaertner2021}, as shown 
in \figref{fig:reference-frames}, right. Each sphere defines a reference frame 
$\notationFrame{S}_{i}$, and it is parametrized by a fixed translation 
$\notationVectorFrame{t}{B}{\notationFrame{BS}_{i}}$ and a radius $\notationScalar{r}{i}$.

Our final formulation computes the optimal acceleration of the body 
${_B}\mathbf{a}$ given the RMPs summarized in \tabref{tab:motion-policies} as 
the following optimization:
%%%%%%%%%%%%%%%%%%%%%%%%%%%%%%%%%%%%%%%%%%%%%%%%%%%%%%%%%%%%%%%%%%%%%%%%%%%%%%%%
% RMP formulation
\begin{equation}
\label{eq:rmp-formulation-detailed}
\argmin_{{_\notationFrame{B}}\mathbf{a}}{
	\displaystyle 
	% Sum of base RMPs
	\underbrace{
	\sum_{ \{{_\notationFrame{B}}\mathbf{f},\, \mathbf{M}\}}{
		\lVert {_\notationFrame{B}}\mathbf{f} - 
		{_\notationFrame{B}}\mathbf{a} 
		\rVert^{2}_{\mathbf{M}}}}_{\text{RMPs applied to body}}
	+
	% Sum of obstacle RMPs
	\underbrace{
	\sum_{\mathcal{S}}{ \ 
	\sum_{ \{{_\notationFrame{S}}\mathbf{f},\, \mathbf{M}\}}{
	\lVert {_\notationFrame{S}}\mathbf{f} - 
	\mathbf{J}_{\notationFrame{SB}}\, {_\notationFrame{B}}\mathbf{a} 
	\rVert^{2}_{\mathbf{M}}}}}_{\text{RMPs applied to collision spheres}}
	}.
\end{equation}

In \eqref{eq:rmp-formulation-detailed}, we expand the original formulation in 
\eqref{eq:rmp-formulation} 
to explicitly state where each motion policy is being applied. This can be either the body frame~$\notationFrame{B}$ or the collision spheres~$\mathcal{S}$. When applying RMPs to collision spheres, they only affect the center point, and therefore the 
accelerations only have translational components.
Hence, the Jacobian --or \emph{pullback}-- $\notationMatrixFrame{J}{}{\notationFrame{SB}}$ is required 
to map the effect of the desired acceleration of the body 
$\notationVectorFrame{a}{B}{}$ to the corresponding collision sphere. 

The resulting acceleration $\notationVectorFrame{a}{B}{}$ is time-integrated at 
the controller 
rate (\SI{10}{\hertz}) to obtain a body twist 
$\notationVectorFrame{v}{B}{}$ which is then passed to the whole-body 
controller (WBC). For real experiments we use a WBC, developed by collaborators 
and based on Lee \etal~\cite{Lee2020}, that generates a valid walking motion 
given a desired velocity using a learned gait model, which is robust over 
uneven terrains. Our simulation experiments use the default model-based WBC 
from the manufacturer.

%%%%%%%%%%%%%%%%%%%%%%%%%%%%%%%%%%%%%%%%%%%%%%%%%%%%%%%%%%%%%%%%%%%%%%%%%%%%%%%%
% Motion policies table
\begin{table*}[t]
	\vspace{1mm}
	\caption{\small{Main Riemannian Motion Policies (RMP) used for the reactive 
	controller formulation.}}
	\centering
	\begin{threeparttable}
	{\footnotesize
	\begin{tabular}{p{0.08\linewidth}
					p{0.125\linewidth}
					p{0.125\linewidth}
					p{0.125\linewidth}
					p{0.125\linewidth}
					p{0.125\linewidth}
					p{0.125\linewidth}
				}
		\thickhline
		\textbf{RMP} & 
			GDF-based Goal Reaching & 
			Obstacle-free Goal Reaching & 
			Obstacle Avoidance & 
			Heading Correction & 
			Damping & 
			Regularization \\
		\hline	
		% Description row
		\textbf{Description}  & 
			Uses the GDF to reach the goal &
			Minimizes the position and orientation error assuming free space& 
			Uses the SDF to push the robot away from obstacles &
			Aligns the robot with the actual translation-only velocity vector &
			Damps the output acceleration depending\,\,on the current velocity &
			Smooths the output acceleration using the last acceleration 
			command\\
		% Collision spheres
		Applied to   & 
			Body $\notationFrame{B}$ & 
			Body $\notationFrame{B}$ & 
			Collision sphere $\notationFrame{S}_i$ &
			Body $\notationFrame{B}$ &
			Body $\notationFrame{B}$ &
			Body $\notationFrame{B}$ \\
		\hline
		% acceleration row
		\textbf{Acceleration field $\notatVector{f}$} & 
			$ -k\,\nabla 
			f_{\text{GDF}}(\notationVectorFrame{t}{M}{\notationFrame{MB}})$ & 
			{$ k\,\Logmap{\notationMatrixFrame{T}{}{\notationFrame{IB}}^{-1} 
			\notationMatrixFrame{T}{}{\notationFrame{IG}}}$\tnote{$\star$}}& 
			$ - k\frac{\nabla 
			f_{\text{SDF}}(\notationVectorFrame{t}{M}{\notationFrame{MB}}) 
			}{(f_{\text{SDF}}(\notationVectorFrame{t}{M}{\notationFrame{MB}}) - 
			\notationScalar{r}{i})}$ &
			$ k\,\text{atan2}(\notationScalarFrame{v_y}{B}{}, 
			               \notationScalarFrame{v_x}{B}{}) 
			$ &
			$-k\, \notationVectorFrame{v}{B}{}$ &
			$    \notationVectorFrame{a}{B}{t-1}$\tnote{$\dagger$}  \\
%		Parameters &
%			Gain $k$ &
%			Gain $k$ &
%			Gain $k$, {{sphere radius $r_i$}} &
%			Gain $k$ &
%			Gain $k$ &
%			- \\
		Components affected &
			Only\,\,translation, $\notationScalarFrame{f}{B}{\theta} = 0$ &
			Translation\,and orientation &
			Only\,\,translation, $\notationScalarFrame{f}{S}{\theta} = 0$ &
			Only\,\,orientation, $\notationScalarFrame{f}{B}{x} = 
			\notationScalarFrame{f}{B}{y} = 0$ &
			Translation\,\,and orientation &
			Translation\,\,and orientation\\
		\hline
		
		% Metric
		\textbf{Metric $\notatMatrix{M}$} & 
			$\sigma(d, d_{c})\, \notationMatrix{I}{3\times3}$ 
			\tnote{$\ddagger$}&
			$\sigma(d, d_{c})\, \notationMatrix{I}{3\times3}$ &
			$\sigma(d, d_{c})\, \notationMatrix{I}{2\times2}$ &
			$\sigma(d, d_{c})\, \notationMatrix{I}{3\times3}$ &
			$\sigma(d, d_{c})\, \notationMatrix{I}{3\times3}$ &
			$s\,\notationMatrix{I}{3\times3}$ \\
%		Parameters &
%			Critical distance $d_{c}$ &
%			Critical distance $d_{c}$ &
%			Critical distance $d_{c}$ &
%			Critical distance $d_{c}$ &
%			Critical distance $d_{c}$ &
%			Scaling $s$ \\
		When Enabled? &
			Distance to goal $d$ is \textbf{further} than $d_{c}$ &
			Distance to goal $d$ is \textbf{closer} than $d_{c}$ &
			Distance to obstacle $d$ is \textbf{closer} than $d_{c}$&
			Distance to goal $d$ is \textbf{further} than $d_{c}$&
			Always &
			Always \\

		\thickhline
	\end{tabular}
	}
	% Table footnotes
	\begin{tablenotes}\footnotesize
		\item[$\star$] $\Logmap{\cdot}$ is the logarithm map of 
		$\SEtwo$~\cite{Sola2018}.
		\item[$\dagger$] $\notationVectorFrame{a}{B}{t-1}$ corresponds to the 
		acceleration computed in the previous iteration.
		\item [$\ddagger$] The function $\sigma(d)$ is implemented as a 
		logistic/inverse logistic function that becomes 1 when $d$ is 
		closer/farther than $d_{c}$ depending on the case.
	\end{tablenotes}
	\end{threeparttable}
	\label{tab:motion-policies}
\end{table*}

\subsection{Integration with the \vtr{} System}
\label{sec:vtr-integration}

For closed-loop integration with the \vtr{} system, we use the \emph{carrot-on-a-stick} strategy,
which sends a look-ahead goal at a distance $d_{\text{carrot}}$ which ensures 
that the robot moves
continuously along reference trajectory.
Additionally, we implemented status feedback from the 
controller to the \vtr{} system, which not only reports when a goal is 
reached, but also if a goal is unreachable, e.g, when the path is 
blocked. As reported in previous work~\cite{Mattamala2021}, our original 
\vtr{} system achieves less than \SI{20}{\centi\meter} error, and 
\SI{10}{\centi\meter} on average.

%%%%%%%%%%%%%%%%%%%%%%%%%%%%%%%%%%%%%%%%%%%%%%%%%%%%%%%%%%%%%%%%%%%%%%%%%%%%%%%%
\section{Results}
\label{sec:results}

We performed experiments both in simulation and on the ANYmal C100 robot with 
an Intel i7 computer onboard. All the processing was CPU-based.
The robot has been modified from the stock version by replacing the front and 
rear cameras with two RoboSense BPearl LiDARs. The data from the LiDARs was 
used to build local elevation maps using the approach of Fankhauser \etal~\cite{Fankhauser2018}. Additionally, a front wide-angle stereo camera developed by Sevensense Robotics was installed 
on top and was used to feed the \vtr{} system.

The parameters in \tabref{tab:motion-policies} were tuned using line search, 
and validated online in simulation and the real robot. We tested our 
system in a cluttered indoor space, as well as a large-scale 
underground mine. We evaluated the performance of our system both 
quantitatively and qualitatively in the two scenarios, as well as the 
computation time required. Simulation experiments were used to provide a 
qualitative study of the contribution of individual motion policies and compare 
against other baselines.

\subsection{Study 1: Influence of each RMP}
\begin{figure}[h]
	\centering
	\includegraphics[trim={0cm 0cm 0cm 0cm},
	width=0.95\columnwidth]{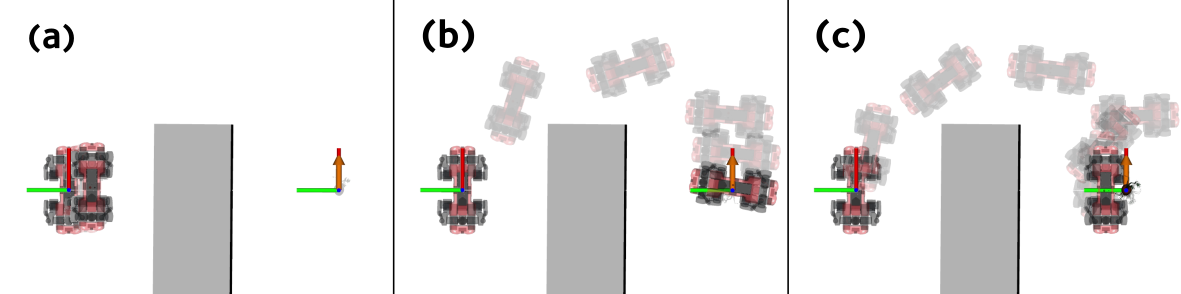}
	\caption{\small{Study 1: Influence of RMPs. 
			\textbf{(a)} \emph{Obstacle-free Goal Reaching} + \emph{Obstacle 
			Avoidance RMPs}.
			\textbf{(b)} \emph{GDF-based Goal Reaching} + \emph{Obstacle 
			Avoidance RMPs}.
			\textbf{(c)} All the RMPs in 
			\tabref{tab:motion-policies}.}}
	\label{fig:sim-experiment}
\end{figure}

In this experiment, we provide a qualitative analysis performed in a simulated environment to understand 
the influence of the different RMPs considered. 
In \figref{fig:sim-experiment}, we show three examples which use different 
combinations of the RMPs.
The robot was given a goal on the other side of the wall with the heading in 
the same direction as the starting pose. 
\figref{fig:sim-experiment}\,(a) shows that by using only \emph{Obstacle-free Goal Reaching} + 
\emph{Obstacle Avoidance} RMPs, the solution falls into a local minimum as the 
goal was at the other side of the wall, though the robot kept a safe distance 
from the wall.

By using the \emph{GDF-based Goal Reaching} and \emph{Obstacle Avoidance} RMPs, 
the robot is able to go around the wall. However, the final orientation of the robot 
does not match the goal, having $\approx$\SI{90}{\degree} error as seen in 
\figref{fig:sim-experiment}\,(b), as the GDF-based RMP does not have an 
orientation component.

In \figref{fig:sim-experiment}\,(c), 
the weighted combination of RMPs via distance-enabled metrics allows the robot 
to go around the obstacle by relying on the GDF when the goal is farther than a distance~$d_c$,
whereas it tries to minimize orientation error using the \emph{Obstacle-free Goal Reaching} RMP 
within a distance~$d_c$ of the goal, where $d_c$ = \SI{1}{\meter} in our experiments.

\subsection{Study 2: Computation time}
\label{sec:computation-time}
\begin{figure}[h]
	\centering
	\includegraphics[trim={0cm 0.5cm 0cm 0cm},
	width=1\columnwidth]{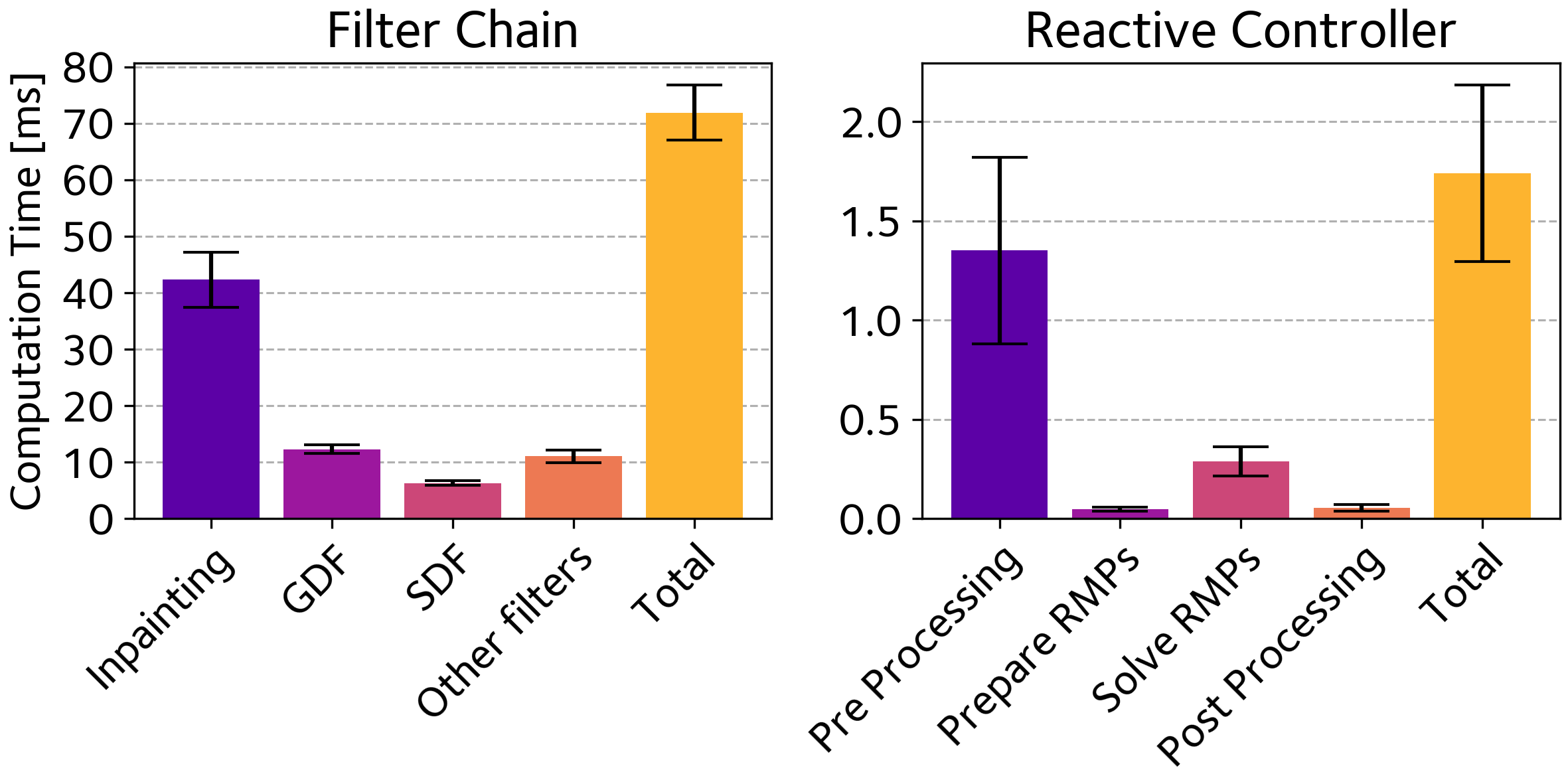}
	\caption{\small{Study 2: Computation time of the filter chain and 
			reactive controller.}}
	\label{fig:computation-time}
\end{figure}

We measured the computation time of the filter chain 
and the controller. We ran our controller on the real robot to collect data for 
about \SI{5}{\minute} during a typical \vtr{} mission.
\figref{fig:computation-time} shows the computation time across the 
whole experiment, obtaining an upper bound of \SI{80}{\milli\second} 
($\approx$\SI{12}{\hertz}) on average. Most of the computation is spent in the 
in-painting operation, followed by the GDF computation which requires about
\SI{13}{\milli\second} on average. Using coarser grid resolutions, smaller 
local maps, or simpler in-painting strategies could help to increase the 
operation frequency if required.

By comparison, the controller has minimum computational cost, requiring 
$<$\SI{2}{\milli\second} on average. Setting up the RMPs and solving the 
optimization problem also has a negligible cost due to the simplicity of the 
approach. This could allow us to operate at higher frequency if we reuse
the last elevation map processed by the filter chain. 

\subsection{Experiment 1: Cluttered Indoor Space}

\begin{figure}[t]
	\vspace{1mm}
	\centering
	\includegraphics[trim={0cm 0cm 0cm 0cm},
	width=0.95\columnwidth]{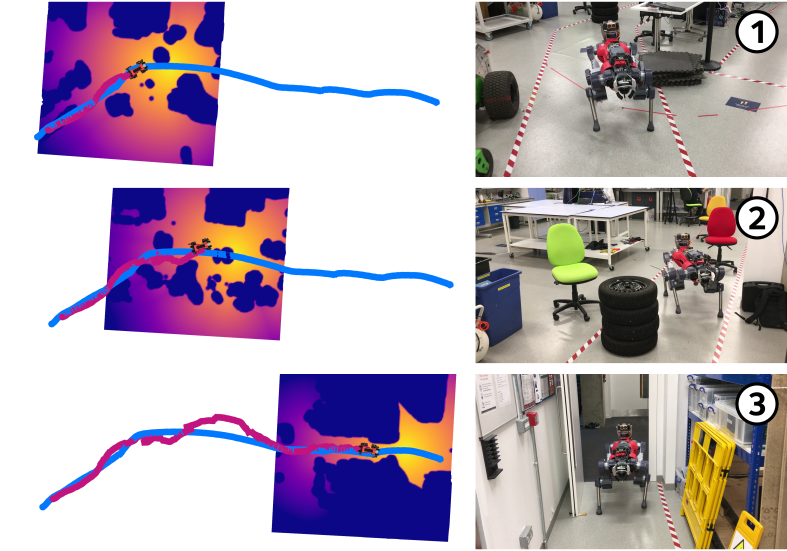}
	\caption{\small{Experiment 1: Indoor Space. Example of a 
			repeat traversal. Blue \bluesquare{} is the original teach path; 
			magenta \magentasquare{} the repeat traversals after introducing 
			obstacles. 
			\textbf{Left:} traversed path as well as the GDF 
			computed online. 
			\textbf{Right:} Footage from the actual traversal.
	}}
	\label{fig:batlab-experiment}
\end{figure}

This experiment was designed to demonstrate and evaluate the performance 
of our \vtr{} system integrated with the reactive controller in a cluttered indoor environment.  
We performed a teach step along a short trajectory (\SI{18}{\meter}) in our lab 
environment.
We placed objects along the route to create an obstacle course to demonstrate 
(1) the performance of our reactive controller in terms of avoiding obstacles, and 
(2) the robustness of the overall system to deal with the degraded appearance 
of the environment introduced due to obstacles.
We ran the system in repeat mode four times (twice walking forward and twice 
backward),
moving the obstacles between each run.
\figref{fig:batlab-experiment} illustrates some of the behaviors emerging after introducing obstacles to 
the reference path.

\begin{figure}[h]
	\vspace{1mm}
	\centering
	\includegraphics[trim={0cm 0cm 0cm 0cm},
	width=\columnwidth]{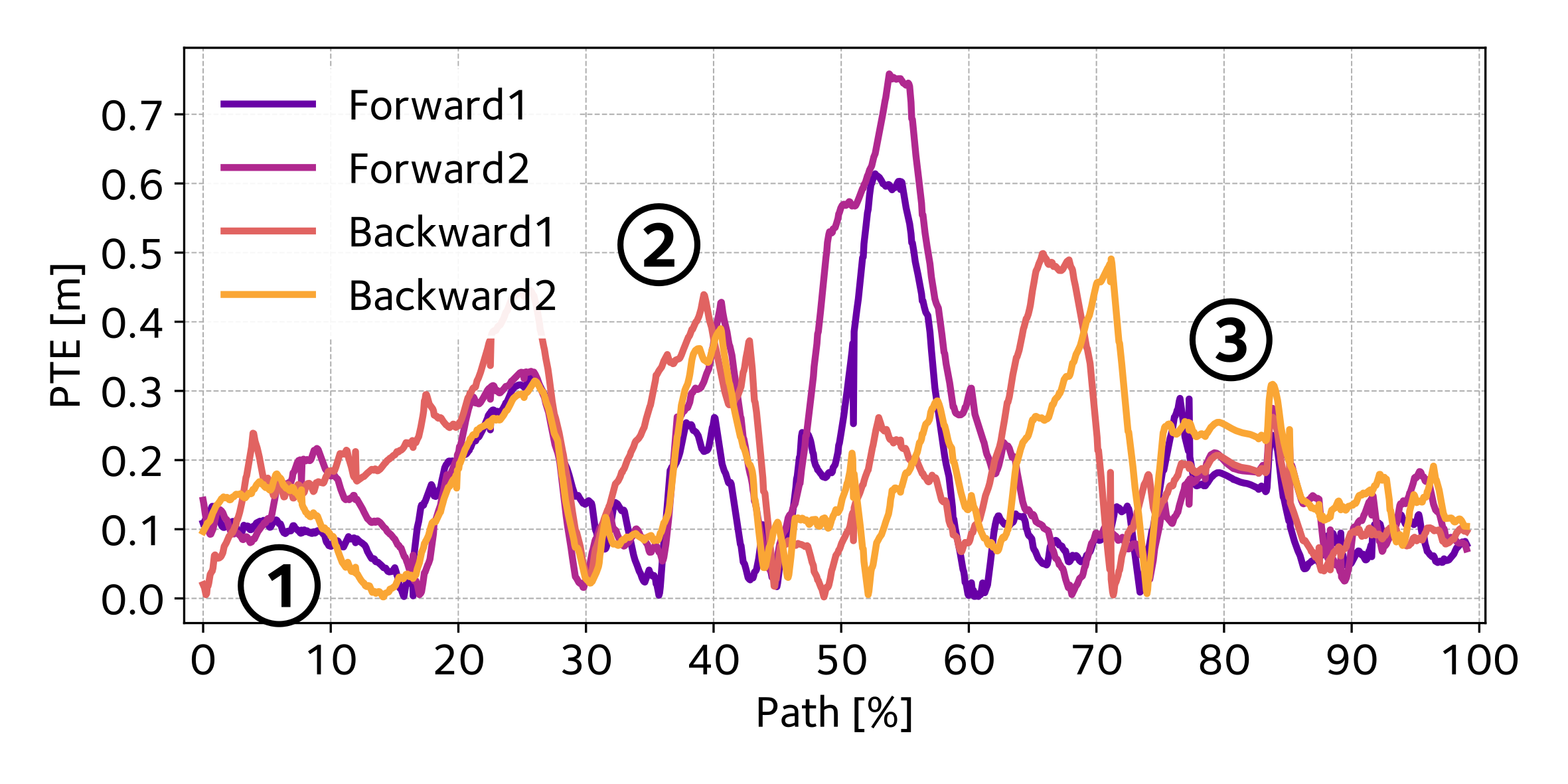}
	\caption{\small{Experiment 1: Path tracking error for the 
			four different traversals in the indoor environment.
	}}
	\label{fig:batlab-experiment-pte}
\end{figure}

We computed the \emph{path tracking error} (PTE) to evaluate the 
tracking performance of \vtr{} systems~\cite{Mattamala2021} shown in 
\figref{fig:batlab-experiment-pte}. Far from obstacles, the system performed 
similar to our previous approach~\cite{Mattamala2021}, but the tracking 
performance decreased near them (indicated with the spikes in the 
errors, numbered \boldcircled{2} and \boldcircled{3}), which is consistent
as the controller forces the robot to deviate from the reference path using the 
cues given by the GDF as well as the obstacle avoidance policies.
In the first run (\emph{Forward1}), we used a carrot distance 
$d_{\text{carrot}} =$ \SI{0.7}{\meter} resulting in an average speed of 
\SI{0.16}{\meter\per\second}. In the other traversals, we used 
$d_{\text{carrot}} =$ \SI{1.5}{\meter}
which allowed the robot to run faster at \SI{0.31}{\meter\per\second} on 
average.

\subsection{Experiment 2: Large Underground Mine}

\begin{figure*}[t]
	\vspace{3mm}
	\centering
	\includegraphics[trim={0cm 0cm 0cm 0cm},
	width=2\columnwidth]{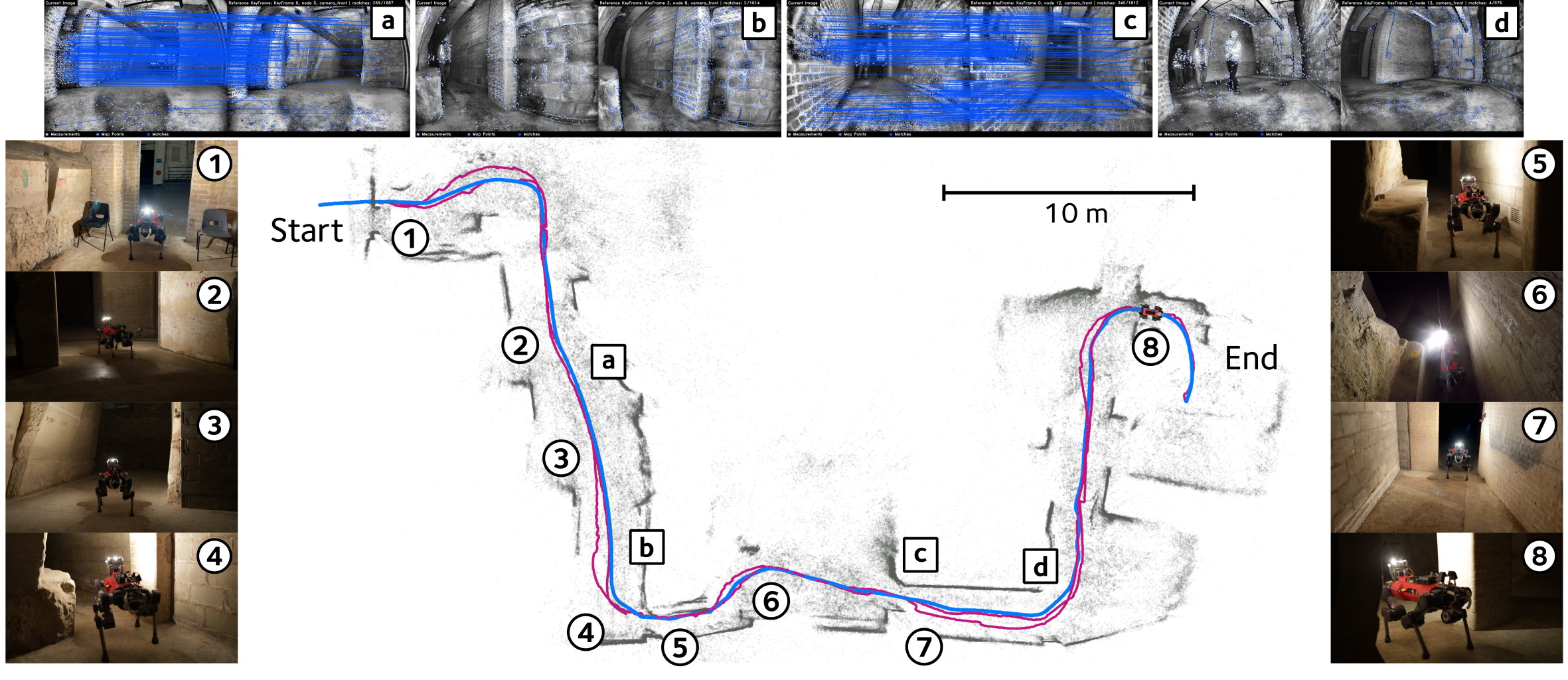}
	\caption{\small{Experiment 2: Underground Mine. 
			Blue~\bluesquare{} \SI{60}{\meter} teach trajectory; 
			magenta~\magentasquare{} two \vtr{}-estimated repeat trajectories 
			(forward and backward). The map in 
			gray~\graysquare{} was not used by the \vtr{} system and 
			is only shown to illustrate the test area.
			The pictures on the sides depict the robot executing the mission, 
			while examples of visual tracking are shown on top.
		}}
	\label{fig:corsham-experiment}
\end{figure*}

Our second experiment was executed in an unlit decommissioned underground 
mine in Wiltshire, UK. We teleoperated the robot along a \SI{60}{\meter} long 
teach path which passed through narrow passages close to walls and large areas 
that challenged the reliability of stereo vision.
The robot carried onboard lights and we enhanced the images using CLAHE 
equalization~\cite{Zuiderveld1994} to deal with the poor lighting conditions, 
as shown in \figref{fig:corsham-experiment} and the attached video.

In repeat mode, we had the robot traverse the path twice -- forward as well as 
backward.
\figref{fig:corsham-experiment} shows the resulting trajectory traversed by the robot (center), as well as some 
examples of visual tracking from the \vtr{} system 
(\boldsquared{a}-\boldsquared{d} on the top).
We also illustrate some of the challenging areas that the robot traversed during the repeat phase, and
highlight these areas along the trajectory (\figref{fig:corsham-experiment}, 
\boldcircled{1}\,--\,\boldcircled{8}\, on the sides).
The robot also safely navigated narrow corridors with a width of 60\,cm with 
the opening being only slightly larger (<\,\SI{30}{\centi\meter})
than the robot body's width, as shown in \boldcircled{6}.

\begin{figure}[h]
	\centering
	\includegraphics[trim={0cm 0.5cm 0cm 0cm},
	width=\columnwidth]{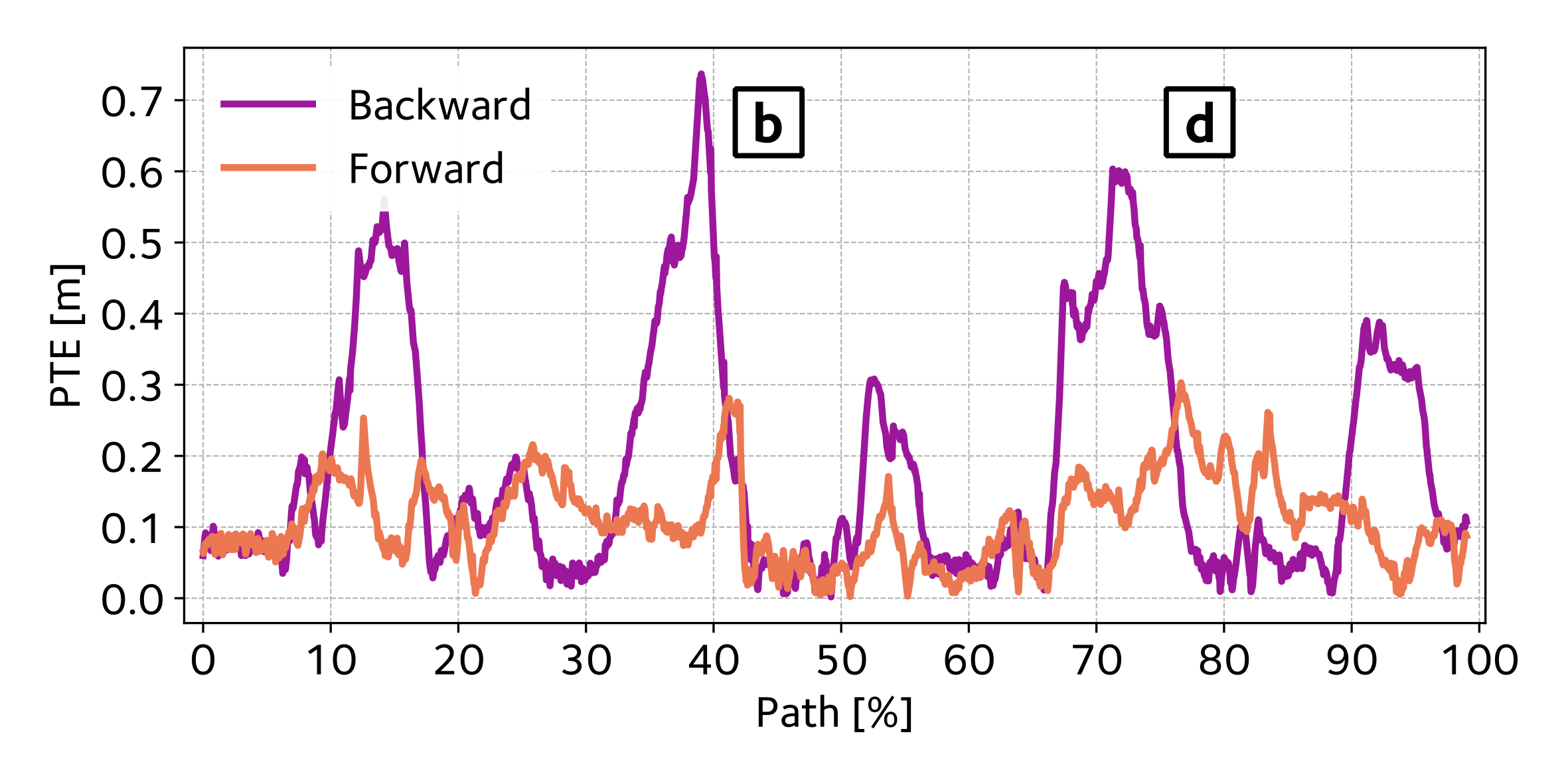}
	\caption{\small{Experiment 2: Path tracking error obtained in the 
	underground mine for forward and backward traversals.}}
	\label{fig:corsham-pte}
\end{figure}

As for the previous experiment, we also show the PTE in 
\figref{fig:corsham-pte}.
The robot achieved better tracking when traversing in the 
same direction (orange \orangesquare{}) than the reverse traversal (purple 
\purplesquare{});
this is consistent with previous studies of the optimal placement of cameras 
for visual odometry~\cite{Peretroukhin2014}. 
The most significant deviations with respect to the reference path were when 
rounding corners when walking backward, see locations \boldsquared{b} and 
\boldsquared{d} in \figref{fig:corsham-experiment} and 
\figref{fig:corsham-pte}. 
Here, visual tracking failed causing the system to fall on
the propagated estimate using the robot's odometry until recovery.
Despite the loss of visual tracking, the robot traversed in both runs 
without collision even in narrow corridors at
\SI{0.3}{\meter\per\second} on average.

\subsection{Experiment 3: Simulation Baselines}
\begin{figure}[h]
	\centering
	\includegraphics[trim={0cm 0.5cm 0cm 0cm},
	width=\columnwidth]{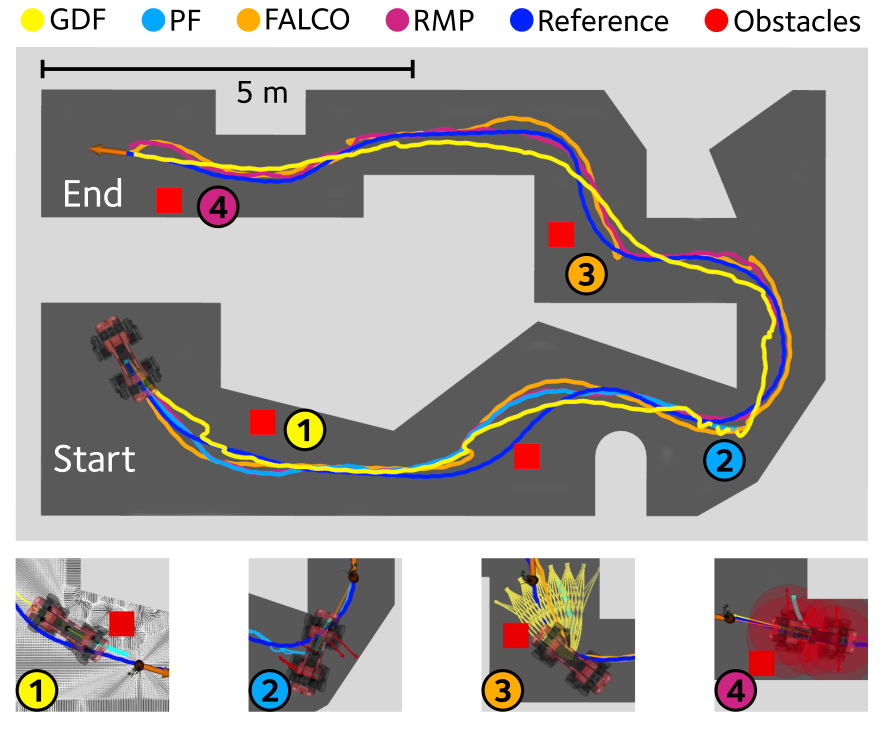}
	\caption{\small{Experiment 3: Representative trajectories tested in 
	simulation for different controllers. The numbered 
	captions illustrate some failure cases observed with each method. Please 
	refer to the text and complementary video for more details.
	}}
	\label{fig:baselines}
\end{figure}

As a final experiment, we showed the advantages of our approach compared to 
other local planners/controllers in a simulated environment. We built the setup 
shown in \figref{fig:baselines}, recorded a collision-free reference path in 
free space, and then we introduced artificial obstacles to force the robot to 
deviate from that path, similarly to Experiment 1. To avoid the overhead of 
running our full \vtr{} system, we only implemented the 
carrot-on-a-stick approach using the reference path to send look-ahead goals 
to the controllers. The baseline methods used were:

\begin{itemize}
	\item \emph{FALCO-based local planner}: We adapted the planner released by 
	CMU 
	Exploration\footnote{\url{https://www.cmu-exploration.com/tare-planner}}, 
	based on collision-checking and scoring of precomputed 	
	trajectories~\cite{Zhang2020}, and tracking the best-scored path with a P 
	controller.
	\item \emph{Potential Field (PF)}: We made our RMP controller behave like a 
	PF by using \emph{only} the \emph{Obstacle-free Goal Reaching} + 
	\emph{Obstacle 	Avoidance} RMPs, with fixed metrics.
	\item \emph{GDF-only}: Similarly, we only kept enabled the \emph{GDF-based} 
	and \emph{Obstacle-free Goal Reaching} RMPs.
\end{itemize}

The controllers were tuned using line search, and executed 10 times each; we 
recorded the trajectories and number of collisions. \figref{fig:baselines} 
shows representative trajectories obtained with each method, as well as failure 
cases.

Our results show that, while most of the methods are able to reach the final 
goal, all the controllers except our proposed method hit the walls more 
than 10 times. The GDF controller was able to reach the final position but 
repeatedly bumped into the walls due to the lack of local awareness 
\boldcircled{1}. PF managed to traverse the first third of the trajectory 
without collisions, but in 5 out of 10 trials it got stuck in a corner 
\boldcircled{2} due to local minima. The precomputed trajectories of FALCO 
(yellow \yellowsquare{}) allowed it to avoid most of the obstacles smoothly, 
but the robot was unable to pass through sharp curves such as \boldcircled{3}, 
since the planner was unable to find free 
paths, or the P controller for path tracking was not able to react 
appropriately.

In contrast, our proposed RMP controller showed local awareness as PFs do, but 
the metrics (red circles \redsquare{} in \figref{fig:baselines}) allowed it to 
leverage the fields dynamically, avoiding most of the obstacles. We only 
observed 3 minor bumps near the goal position \boldcircled{4}, which are shown 
in the attached video.

%%%%%%%%%%%%%%%%%%%%%%%%%%%%%%%%%%%%%%%%%%%%%%%%%%%%%%%%%%%%%%%%%%%%%%%%%%%%%%%%
\section{Conclusions}
\label{sec:conclusions}
In this work we presented a locally reactive controller coupled with our \vtr{} 
navigation system that allowed safe navigation in challenging environments. Our 
approach is based on computing efficient vector field representations from a 
local map, which are used to compute acceleration fields. We combined them 
using a Riemannian Motion Policies controller to generate a twist command at 
\SI{10}{\hertz}. We deployed our system on an ANYmal robot and showed that our 
locally reactive controller allowed the robot to traverse cluttered paths 
taught without obstacles. We also deployed the system in an underground mine, 
demonstrating safe navigation through challenging rooms and corridors, walking 
at \SI{0.30}{\meter\per\second} on average.

%For the next steps, we aim to extend our system to autonomously navigate in 3D 
%environments using perceptive whole-body controllers, as well as dealing with 
%drastic appearance changes using learning-based methods.

%%%%%%%%%%%%%%%%%%%%%%%%%%%%%%%%%%%%%%%%%%%%%%%%%%%%%%%%%%%%%%%%%%%%%%%%%%%%%%%%

\appendix[Frame transformations of acceleration fields]
\label{sec:appendix}
The \emph{GDF-based Goal Reaching} and \emph{Obstacle 
Avoidance RMPs} require to evaluate the gradients of SDF and GDF to 
generate acceleration fields in the body frame $\notationFrame{B}$.
We determine the pose of the body in the map frame $\notationFrame{M}$ 
as (see \figref{fig:reference-frames}):  
\begin{equation}
\notationMatrixFrame{T}{}{\notationFrame{MB}} = 
   \notationMatrixFrame{T}{}{\notationFrame{IM}}^{-1} 
   \notationMatrixFrame{T}{}{\notationFrame{IB}}
\end{equation}
This is used to evaluate the $\nabla f_{\text{SDF}}$ and $\nabla 
f_{\text{GDF}}$ at the desired location 
$\notationVectorFrame{t}{M}{\notationFrame{MB}}$ (translation components of 
$\notationMatrixFrame{T}{}{\notationFrame{MB}}$). 
However, as the gradients are in $\notationFrame{M}$, we 
need to transform them to frame $\notationFrame{B}$.

The $\SEtwo$ adjoint of $\notationMatrixFrame{T}{}{\notationFrame{MB}}$, 
denoted by $\Adjop{\notationMatrixFrame{T}{}{\notationFrame{MB}}}$, maps 
vectors defined 
in $\notationFrame{B}$ to the frame $\notationFrame{M}$ ~\cite{Sola2018}. Its 
inverse does the opposite, allowing us to transform vectors defined in 
$\notationFrame{M}$ to $\notationFrame{B}$:
\begin{equation}
\Adjop{\notationMatrixFrame{T}{}{\notationFrame{MB}}}^{-1} = 
\Adjop{\notationMatrixFrame{T}{}{\notationFrame{MB}}^{-1}} = 
\Adjop{\notationMatrixFrame{T}{}{\notationFrame{BM}}}.
\end{equation}

The adjoint is given by:
\begin{equation}
\label{eq:adjoint}
\Adjop{\notationMatrixFrame{T}{}{\notationFrame{BM}}} = 
\left[
\begin{smallmatrix}
	\notationMatrixFrame{R}{}{\notationFrame{BM}}  &  
	\hatop{\notatVector{1}}\, \notationVectorFrame{t}{M}{\notationFrame{BM}} 
	\\
	0_{1\times2} &   1
\end{smallmatrix}
\right]
\end{equation}
where the hat operator $\hatop{(\cdot)}$ builds a skew-symmetric matrix from 
the vector, and $\notationMatrixFrame{R}{}{\notationFrame{BM}}$ and 
$\notationVectorFrame{t}{M}{\notationFrame{BM}}$ are the rotation and 
translation components of $\notationMatrixFrame{T}{}{\notationFrame{BM}}$, 
respectively.

The gradient $\nabla f_{\text{GDF}}$ is a vector in $\Rn{2}$ so we 
can add an extra zero for the angular component to define $\nabla 
f'_{\text{SDF}} \in \Rn{3}$. We then use the
adjoint to transform the gradients to $\notationFrame{B}$:
\begin{equation}
\label{eq:GDF-transform}
{_\notationFrame{B}}\nabla f'_{\text{GDF}} = 
\Adjop{\notationMatrixFrame{T}{}{\notationFrame{BM}}}
\nabla f'_{\text{GDF}}.
\end{equation}

We expand \eqref{eq:GDF-transform} using \eqref{eq:adjoint} to 
obtain:
\begin{equation}
\label{eq:SDF-transform-simple}
{_\notationFrame{B}}\nabla f_{\text{GDF}} = 
\notationMatrixFrame{R}{}{\notationFrame{BM}}
\nabla f_{\text{GDF}}
\end{equation}
which basically rotates the gradients to match the frame 
$\notationFrame{B}$.

Evaluating the SDF is similar but we add a transformation 
defined at the center of the collision sphere at $\notationFrame{S}_i$:
\begin{equation}
\notationMatrixFrame{T}{}{\notationFrame{MS}_i} = 
\notationMatrixFrame{T}{}{\notationFrame{IM}}^{-1} \, 
\notationMatrixFrame{T}{}{\notationFrame{IB}} \, 
\notationMatrixFrame{T}{}{}(\notationVectorFrame{t}{B}{\notationFrame{BS}_{i}})
\end{equation}
with 
$\notationMatrixFrame{T}{}{}(\notationVectorFrame{t}{B}{\notationFrame{BS}_{i}})$
a transformation matrix built from the position of the sphere in the body 
frame.

\balance
%\bibliographystyle{IEEEtran}
%\bibliography{library}
\bibliographystyle{./IEEEtran}
\bibliography{./IEEEabrv,./library}

\end{document}

%% file: definitions_drs.tex
\usepackage[utf8]{inputenc}
\usepackage{graphicx} % for pdf, bitmapped graphics files
\usepackage{amsmath} % assumes amsmath package installed
\usepackage{amssymb}  % assumes amsmath package installed
\usepackage[protrusion=false]{microtype}
\usepackage{xcolor}
\usepackage{tensor}
\usepackage{mathtools}
\usepackage{makecell}
\usepackage{lipsum}
\usepackage{booktabs}
\usepackage{mathrsfs}
\usepackage[colorinlistoftodos]{todonotes}
\usepackage{multirow}
\usepackage{subcaption}
\usepackage[linesnumbered,ruled,vlined]{algorithm2e} % for algorithm environment
\usepackage{tabularx}
\usepackage{balance}
\usepackage{threeparttable}
\usepackage{tablefootnote}

\newcolumntype{Y}{>{\centering\arraybackslash}X}
\usepackage{siunitx}
\usepackage[hidelinks]{hyperref} %

\SetKwInput{KwInput}{Input} % Set the Input
\SetKwInput{KwOutput}{Output} % set the Output

\makeatletter
\newcommand{\thickhline}{%
    \noalign {\ifnum 0=`}\fi \hrule height 1pt
    \futurelet \reserved@a \@xhline
}
\newcolumntype{"}{@{\hskip\tabcolsep\vrule width 1pt\hskip\tabcolsep}}
\makeatother

\usepackage[T1]{fontenc}
\usepackage{textcomp}

\def\secref#1{Sec.~\ref{#1}}
\def\figref#1{Fig.~\ref{#1}}
\def\tabref#1{Tab.~\ref{#1}}
\def\eqref#1{Eq.~(\ref{#1})}

%% file: definitions_custom.tex
\usepackage{amsmath}
\usepackage{amssymb}
\usepackage{amsfonts}
\usepackage{dsfont}
\usepackage{pifont}% http://ctan.org/pkg/pifont

\usepackage{etoolbox}	% robustify command
\usepackage{mathtools} % smallmatrix

\newcommand{\etal}{\emph{et al.}\ }

\newcommand{\argmin}{\operatornamewithlimits{arg\,min}}
\newcommand{\Rn}[1]{\mathbb{R}^{#1}}

\newcommand{\subarrow}[1]{
	\mathord{
		\renewcommand{\arraystretch}{0}
		\begin{array}[t]{@{}c@{}l@{}}
			#1\\[2pt]
			\hspace{-2pt}\scriptstyle\longrightarrow
		\end{array}
		\kern\scriptspace
	}
}
\newcommand{\notationFrame}[1]{{\mathtt{#1}}}
\newcommand{\notatFrame}[1]{\subarrow{\mathcal{F}}{}_{\mathtt{#1}}}

\newcommand{\notatMatrix}[1]{\boldsymbol{\mathrm{#1}}}
\newcommand{\notatVector}[1]{\boldsymbol{\mathrm{#1}}}
\newcommand{\notatScalar}[1]{{#1}}
\newcommand{\notatHomog}[1]{\boldsymbol{{#1}}}

\newcommand{\notationMatrix}[2]{\boldsymbol{\mathrm{#1}}_{#2}}
\newcommand{\notationVector}[2]{\boldsymbol{\mathrm{#1}}_{#2}}
\newcommand{\notationScalar}[2]{{#1}_{#2}}
\newcommand{\notationHomog}[2]{\boldsymbol{{#1}}_{#2}}

\newcommand{\notationMatrixFrame}[3]{{_\mathtt{#2}}\boldsymbol{\mathrm{#1}}_{#3}}
\newcommand{\notationVectorFrame}[3]{{_\mathtt{#2}} \boldsymbol{ 
\mathrm{#1}}_{#3}}
\newcommand{\notationScalarFrame}[3]{{_\mathtt{#2}}{#1}_{#3}}
\newcommand{\notationHomogFrame}[3]{{_\mathtt{#2}}\boldsymbol{{#1}}_{#3}}

\robustify{\notatFrame}
\robustify{\notatMatrix}
\robustify{\notatVector}
\robustify{\notatScalar}
\robustify{\notatHomog}
\robustify{\notationMatrix}
\robustify{\notationVector}
\robustify{\notationScalar}
\robustify{\notationHomog}
\robustify{\notationMatrixFrame}
\robustify{\notationVectorFrame}
\robustify{\notationScalarFrame}
\robustify{\notationHomogFrame}

\newcommand{\hatop}[1]{#1^{\wedge}}

\newcommand{\SEtwo}{\mathrm{SE(2)}}

\newcommand{\Logmap}[1]{\mathrm{Log}\left(#1\right)}

\newcommand{\Adjop}[1]{\mathrm{Ad}\left(#1\right)}

% ...
\makeatletter
\renewcommand*\env@matrix[1][\arraystretch]{%
	\edef\arraystretch{#1}%
	\hskip -\arraycolsep
	\let\@ifnextchar\new@ifnextchar
	\array{*\c@MaxMatrixCols c}}
\makeatother

\usepackage{color}
\usepackage{colortbl}
\definecolor{ColorLightCyan}{rgb}{0.88,1,1}
\definecolor{ColorLightTurquoise}{rgb}{0.5, 1, 0.8}
\definecolor{ColorOrange}{rgb}{1.0, 0.7, 0.0}

\newcommand{\vtr}{{VT\&R}}

\newcommand{\magentasquare}{{\textcolor{magenta}{$\blacksquare$}}}
\newcommand{\bluesquare}{{\textcolor{blue}{$\blacksquare$}}}

\newcommand{\graysquare}{{\textcolor{gray}{$\blacksquare$}}}
\newcommand{\orangesquare}{{\textcolor{orange}{$\blacksquare$}}}
\newcommand{\yellowsquare}{{\textcolor{yellow}{$\blacksquare$}}}
\newcommand{\redsquare}{{\textcolor{red}{$\blacksquare$}}}
\newcommand{\purplesquare}{{\textcolor{purple}{$\blacksquare$}}}

\usepackage{tikz}
\newcommand*\notecircle[1]{\tikz[baseline=(char.base)]{
		\node[shape=circle,draw,inner sep=1pt, thick] (char) 
		{\footnotesize{#1}};}}
\newcommand{\boldcircled}[1]{\notecircle{\textbf{#1}}}

\newcommand*\notesquared[1]{\tikz[baseline=(char.base)]{
		\node[shape=rectangle,draw,inner sep=2pt, thick] (char) 
		{\footnotesize{#1}};}}
\newcommand{\boldsquared}[1]{\notesquared{\textbf{#1}}}

%% file: paper.bbl
\begin{thebibliography}{10}
\providecommand{\url}[1]{#1}
\csname url@samestyle\endcsname
\providecommand{\newblock}{\relax}
\providecommand{\bibinfo}[2]{#2}
\providecommand{\BIBentrySTDinterwordspacing}{\spaceskip=0pt\relax}
\providecommand{\BIBentryALTinterwordstretchfactor}{4}
\providecommand{\BIBentryALTinterwordspacing}{\spaceskip=\fontdimen2\font plus
\BIBentryALTinterwordstretchfactor\fontdimen3\font minus
  \fontdimen4\font\relax}
\providecommand{\BIBforeignlanguage}[2]{{%
\expandafter\ifx\csname l@#1\endcsname\relax
\typeout{** WARNING: IEEEtran.bst: No hyphenation pattern has been}%
\typeout{** loaded for the language `#1'. Using the pattern for}%
\typeout{** the default language instead.}%
\else
\language=\csname l@#1\endcsname
\fi
#2}}
\providecommand{\BIBdecl}{\relax}
\BIBdecl

\bibitem{Furgale2010}
P.~Furgale and T.~D. Barfoot, ``{Visual teach and repeat for long-range rover
  autonomy},'' \emph{Journal of Field Robotics}, vol.~27, no.~5, pp. 534--560,
  2010.

\bibitem{Fehr2018}
M.~Fehr, T.~Schneider, and R.~Siegwart, ``{Visual-Inertial Teach and Repeat
  Powered by Google Tango},'' in \emph{IEEE/RSJ Intl. Conf. on Intelligent
  Robots and Systems (IROS)}, 2018.

\bibitem{Churchill2012}
W.~Churchill and P.~Newman, ``{Practice makes perfect? Managing and Leveraging
  Visual Experiences for Lifelong Navigation},'' in \emph{IEEE Intl. Conf. on
  Robotics and Automation (ICRA)}, 2012.

\bibitem{Ratliff2018}
N.~D. Ratliff, J.~Issac, D.~Kappler, S.~Birchfield, and D.~Fox, ``{Riemannian
  Motion Policies},'' \emph{CoRR}, 2018.

\bibitem{Mattamala2021}
M.~Mattamala, M.~Ramezani, M.~Camurri, and M.~Fallon, ``{Learning Camera
  Performance Models for Active Multi-Camera Visual Teach and Repeat},'' in
  \emph{IEEE Intl. Conf. on Robotics and Automation (ICRA)}, 2021.

\bibitem{Kappler2018}
D.~Kappler \emph{et~al.}, ``{Real-time Perception meets Reactive Motion
  Generation},'' \emph{IEEE Robotics and Automation Letters}, vol.~3, no.~3,
  pp. 1864--1871, 2018.

\bibitem{Buchanan2021}
R.~Buchanan, L.~Wellhausen, M.~Bjelonic, T.~Bandyopadhyay, N.~Kottege, and
  M.~Hutter, ``{Perceptive Whole-body Planning for Multilegged Robots in
  Confined Spaces},'' \emph{Journal of Field Robotics}, vol.~38, no.~1, pp.
  68--84, 2021.

\bibitem{Gaertner2021}
M.~Gaertner, M.~Bjelonic, F.~Farshidian, and M.~Hutter, ``{Collision-Free MPC
  for Legged Robots in Static and Dynamic Scenes},'' in \emph{IEEE Intl. Conf.
  on Robotics and Automation (ICRA)}, 2021.

\bibitem{Hoeller2021}
D.~Hoeller, L.~Wellhausen, F.~Farshidian, and M.~Hutter, ``{Learning a State
  Representation and Navigation in Cluttered and Dynamic Environments},''
  \emph{IEEE Robotics and Automation Letters}, vol.~6, no.~3, pp. 5081--5088,
  2021.

\bibitem{Kim2020}
D.~Kim \emph{et~al.}, ``{Vision Aided Dynamic Exploration of Unstructured
  Terrain with a Small-Scale Quadruped Robot},'' in \emph{IEEE Intl. Conf. on
  Robotics and Automation (ICRA)}, 2020.

\bibitem{Vasilopoulos2020}
V.~Vasilopoulos \emph{et~al.}, ``{Reactive Semantic Planning in Unexplored
  Semantic Environments Using Deep Perceptual Feedback},'' \emph{IEEE Robotics
  and Automation Letters}, vol.~5, no.~3, pp. 4455--4462, 2020.

\bibitem{Bista2021}
S.~R. Bista, B.~Ward, and P.~Corke, ``{Image-Based Indoor Topological
  Navigation with Collision Avoidance for Resource-Constrained Mobile
  Robots},'' \emph{Journal of Intelligent and Robotic Systems}, vol. 102,
  no.~3, pp. 1--24, 2021.

\bibitem{Bradley2015}
D.~M. Bradley \emph{et~al.}, ``{Scene understanding for a high-mobility walking
  robot},'' in \emph{IEEE/RSJ Intl. Conf. on Intelligent Robots and Systems
  (IROS)}, 2015.

\bibitem{Krusi2017}
P.~Kr{\"{u}}si, P.~Furgale, M.~Bosse, and R.~Siegwart, ``{Driving on Point
  Clouds: Motion Planning, Trajectory Optimization, and Terrain Assessment in
  Generic Nonplanar Environments},'' \emph{Journal of Field Robotics}, vol.~34,
  no.~5, pp. 940--984, 2017.

\bibitem{Brandao2020}
M.~Brandão, O.~B. Aladag, and I.~Havoutis, ``{GaitMesh: Controller-Aware
  Navigation Meshes for Long-Range Legged Locomotion Planning in Multi-Layered
  Environments},'' \emph{IEEE Robotics and Automation Letters}, vol.~5, no.~2,
  pp. 3596--3603, 2020.

\bibitem{Fankhauser2016GridMapLibrary}
P.~Fankhauser and M.~Hutter, ``{A Universal Grid Map Library: Implementation
  and Use Case for Rough Terrain Navigation},'' in \emph{Robot Operating System
  (ROS) – The Complete Reference}, A.~Koubaa, Ed.\hskip 1em plus 0.5em minus
  0.4em\relax Springer International Publishing, 2016, pp. 99--120.

\bibitem{Nardi2019}
L.~Nardi and C.~Stachniss, ``{Actively Improving Robot Navigation on Different
  Terrains Using Gaussian Process Mixture Models},'' in \emph{IEEE Intl. Conf.
  on Robotics and Automation (ICRA)}, 2019.

\bibitem{Wellhausen2019}
L.~Wellhausen, A.~Dosovitskiy, R.~Ranftl, K.~Walas, C.~Cadena, and M.~Hutter,
  ``{Where Should I Walk? Predicting Terrain Properties From Images Via
  Self-Supervised Learning},'' \emph{IEEE Robotics and Automation Letters},
  vol.~4, no.~2, pp. 1509--1516, 2019.

\bibitem{Oleynikova2016}
H.~Oleynikova, A.~Millane, Z.~Taylor, E.~Galceran, J.~Nieto, and R.~Siegwart,
  ``{Signed Distance Fields: A Natural Representation for Both Mapping and
  Planning},'' in \emph{Robotics: Science and Systems}, 2016.

\bibitem{Mainprice2020}
J.~Mainprice, N.~Ratliff, M.~Toussaint, and S.~Schaal, ``{An Interior Point
  Method Solving Motion Planning Problems with Narrow Passages},'' in
  \emph{IEEE Intl. Conf. on Robot and Human Interactive Communication
  (RO-MAN)}, 2020.

\bibitem{Sethian1996}
J.~A. Sethian, ``{A Fast Marching Level Set Method for Monotonically Advancing
  Fronts},'' \emph{Proceedings of the National Academy of Sciences}, vol.~93,
  no.~4, pp. 1591--1595, 1996.

\bibitem{Crane2017}
K.~Crane, C.~Weischedel, and M.~Wardetzky, ``{The Heat Method for Distance
  Computation},'' \emph{Communications of the ACM}, vol.~60, no.~11, p.
  90–99, 2017.

\bibitem{Valero-Gomez2013}
A.~Valero-Gomez, J.~V. Gomez, S.~Garrido, and L.~Moreno, ``{The Path to
  Efficiency: Fast Marching Method for Safer, more Efficient Mobile Robot
  Trajectories},'' \emph{IEEE Robotics and Automation Magazine}, vol.~20,
  no.~4, pp. 111--120, 2013.

\bibitem{Putz2021}
S.~Pütz, T.~Wiemann, M.~Kleine~Piening, and J.~Hertzberg, ``Continuous
  shortest paths vector field navigation on {3D} triangular meshes for mobile
  robots,'' in \emph{IEEE Intl. Conf. on Robotics and Automation (ICRA)}, 2021.

\bibitem{Ostafew2016}
C.~J. Ostafew, A.~P. Schoellig, and T.~D. Barfoot, ``{Robust Constrained
  Learning-based NMPC enabling reliable mobile robot path tracking},''
  \emph{International Journal of Robotics Research}, vol.~33, no.~1, pp.
  133--152, 2016.

\bibitem{Berczi2017}
L.~P. Berczi and T.~D. Barfoot, ``{Looking High and Low: Learning
  Place-dependent Gaussian Mixture Height Models for Terrain Assessment},'' in
  \emph{IEEE/RSJ Intl. Conf. on Intelligent Robots and Systems (IROS)}, 2017.

\bibitem{Cheng2021}
C.-A. Cheng \emph{et~al.}, ``{RMPflow}: A geometric framework for generation of
  multitask motion policies,'' \emph{IEEE Transactions on Automation Science
  and Engineering}, vol.~18, no.~3, pp. 968--987, 2021.

\bibitem{Fankhauser2018}
P.~Fankhauser, M.~Bloesch, and M.~Hutter, ``{Probabilistic Terrain Mapping for
  Mobile Robots With Uncertain Localization},'' \emph{IEEE Robotics and
  Automation Letters}, vol.~3, no.~4, pp. 3019--3026, 2018.

\bibitem{Telea2004}
A.~Telea, ``{An Image Inpainting Technique Based on the Fast Marching
  Method},'' \emph{Journal of Graphics Tools}, vol.~9, no.~1, pp. 23--34, 2004.

\bibitem{Wermelinger2016}
M.~Wermelinger, P.~Fankhauser, R.~Diethelm, P.~A. Krüsi, R.~Siegwart, and
  M.~Hutter, ``\BIBforeignlanguage{en}{{Navigation Planning for Legged Robots
  in Challenging Terrain}},'' in \emph{\BIBforeignlanguage{en}{IEEE/RSJ Intl.
  Conf. on Intelligent Robots and Systems (IROS)}}, 2016.

\bibitem{Felzenszwalb2012}
P.~F. Felzenszwalb and D.~P. Huttenlocher, ``{Distance Transforms of Sampled
  Functions},'' \emph{Theory of Computing}, vol.~8, no.~19, pp. 415--428, 2012.

\bibitem{Peyre2010}
G.~Peyr{\'e}, M.~P{\'e}chaud, R.~Keriven, and L.~D. Cohen, ``{Geodesic Methods
  in Computer Vision and Graphics},'' \emph{{Foundations and Trends in Computer
  Graphics and Vision}}, 2010.

\bibitem{Lee2020}
J.~Lee, J.~Hwangbo, L.~Wellhausen, V.~Koltun, and M.~Hutter, ``{Learning
  quadrupedal locomotion over challenging terrain},'' \emph{Science Robotics},
  vol.~5, no.~47, 2020.

\bibitem{Sola2018}
J.~Sol{\`{a}}, J.~Deray, and D.~Atchuthan, ``{A micro Lie theory for state
  estimation in robotics},'' \emph{arXiv}, 2018.

\bibitem{Zuiderveld1994}
K.~Zuiderveld, ``{Contrast Limited Adaptive Histograph Equalization.}'' in
  \emph{Graphic Gems IV}.\hskip 1em plus 0.5em minus 0.4em\relax Academic Press
  Professional, 1994.

\bibitem{Peretroukhin2014}
V.~Peretroukhin, J.~Kelly, and T.~D. Barfoot, ``{Optimizing Camera Perspective
  for Stereo Visual Odometry},'' in \emph{Canadian Conference on Computer and
  Robot Vision}, 2014.

\bibitem{Zhang2020}
J.~Zhang, C.~Hu, R.~G. Chadha, and S.~Singh, ``{FALCO}: Fast likelihood-based
  collision avoidance with extension to human-guided navigation,''
  \emph{Journal of Field Robotics}, vol.~37, no.~8, pp. 1300--1313, 2020.

\end{thebibliography}
